\documentclass[10pt,journal,compsoc]{IEEEtran}
\usepackage{times}
\usepackage{epsfig}
\usepackage{graphicx}
\usepackage{amsmath}
\usepackage{amssymb}
\usepackage{multirow}
\usepackage{booktabs}
\usepackage{soul}
\usepackage{subcaption}
\usepackage{xspace}
\usepackage{mathrsfs}
\usepackage{soul}
\newcommand{\ie}{{\emph{i.e.}}\xspace}
\newcommand{\eg}{{\emph{e.g.}}\xspace}

\newcommand{\vs}{{\emph{v.s.}}\xspace}
\newcommand{\etal}{{\emph{et al.}}\xspace}
\newcommand{\etc}{{\emph{etc.}}\xspace}
\usepackage{array}
\usepackage{tabu}
\usepackage{color}
\usepackage{xcolor}
\usepackage{flushend}

\ifCLASSOPTIONcompsoc
  \usepackage[nocompress]{cite}
\else
  \usepackage{cite}
\fi
\ifCLASSINFOpdf
\else
\fi
\begin{document}
\title{VLT: Vision-Language Transformer and Query Generation for Referring Segmentation}

\author{Henghui Ding, Chang Liu, Suchen Wang, and Xudong Jiang,~\IEEEmembership{Fellow,~IEEE}% <-this % stops a space
\IEEEcompsocitemizethanks{
% \IEEEcompsocthanksitem Henghui Ding was with School of Electrical and Electronic
% Engineering, Nanyang Technological University, Singapore 639798.\protect\\
% % note need leading \protect in front of \\ to get a newline within \thanks as
% % \\ is fragile and will error, could use \hfil\break instead.
% E-mail: see http://www.michaelshell.org/contact.html
\IEEEcompsocthanksitem Henghui Ding, Chang Liu, Suchen Wang and Xudong Jiang are with the School of Electrical and Electronic
Engineering, Nanyang Technological University (NTU), Singapore 639798. (e-mail: henghuiding@gmail.com; liuc0058@e.ntu.edu.sg; wang.sc@ntu.edu.sg;
exdjiang@ntu.edu.sg)
\IEEEcompsocthanksitem Henghui Ding and Chang Liu are co-first author. 
\IEEEcompsocthanksitem Corresponding author: Henghui Ding.}% <-this % stops an unwanted space
% \thanks{Henghui Ding and Chang Liu are co-first author}
% \thanks{Manuscript received April 19, 2005; revised August 26, 2015.}
}


\markboth{IEEE Transactions on Pattern Analysis and Machine Intelligence}%
{Shell \MakeLowercase{\textit{et al.}}: Bare Demo of IEEEtran.cls for Computer Society Journals}



\IEEEtitleabstractindextext{%
\begin{abstract}
We propose a Vision-Language Transformer (VLT) framework for referring segmentation to facilitate deep interactions among multi-modal information and enhance the holistic understanding to vision-language features. There are different ways to understand the dynamic emphasis of a language expression, especially when interacting with the image. However, the learned queries in existing transformer works are fixed after training, which cannot cope with the randomness and huge diversity of the language expressions. To address this issue, we propose a Query Generation Module, which dynamically produces multiple sets of input-specific queries to represent the diverse comprehensions of language expression. To find the best among these diverse comprehensions, so as to generate a better mask, we propose a Query Balance Module to selectively fuse the corresponding responses of the set of queries. Furthermore, to enhance the model's ability in dealing with diverse language expressions, we consider inter-sample learning to explicitly endow the model with knowledge of understanding different language expressions to the same object. We introduce masked contrastive learning to narrow down the features of different expressions for the same target object while distinguishing the features of different objects. The proposed approach is lightweight and achieves new state-of-the-art referring segmentation results consistently on five datasets.
\end{abstract}

}

\maketitle

\IEEEdisplaynontitleabstractindextext

% \fi
\IEEEpeerreviewmaketitle

% \IEEEraisesectionheading{\section{Introduction}\label{sec:introduction}}

\IEEEraisesectionheading{\section{Introduction}\label{sec:introduction}}

\IEEEPARstart{R}{eferring} segmentation aims at generating a segmentation mask for the target object that are referred by a given query expression in natural language \cite{hu2016segmentation,ding2020phraseclick}.
Referring segmentation is one of the most fundamental while challenging multi-modal tasks, involving both natural language processing and computer vision. It is intensely demanded in many practical applications, \eg, video/image editing, by providing a user-friendly interactive way. Recently, many deep-learning-based methods have arisen in this field and achieved remarkable results. However, great challenges still remain: while the query expression implies the target object by describing its attributes and its relationships with other objects, objects in referring segmentation images relate to each other in a complex manner. Therefore, a holistic understanding of the image and language expression is desired. Another challenge is that the diverse objects/images and the unconstrained language expressions bring a high level of randomness, which requires the modal high generalization ability in understanding different kinds of images and language expressions.

\begin{figure}[t]
   \begin{center}
      \includegraphics[width=1\linewidth]{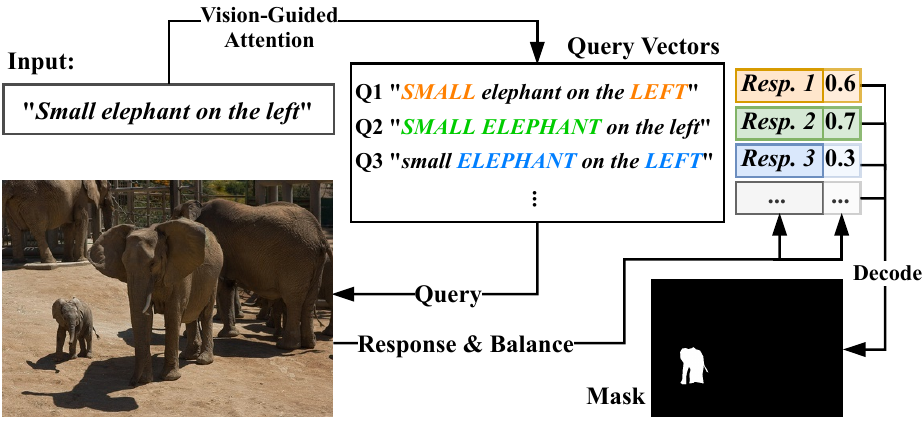}
   \end{center}
   \caption{The proposed method dynamically produces multiple sets of input-specific query vectors to represent the diverse comprehensions of language expression. We use each vector to ``query'' the input image, generating a response to each query. Then the network selectively aggregates the responses to these queries, so that queries that provide better comprehensions are highlighted.}
   \label{fig:fig1}
\end{figure}

Firstly, to address the challenge of complicated correlations in the input image and query expression, we propose to enhance the holistic understanding of multi-modal information by designing a framework with global operations, in which direct interactions are built among all elements, \eg, word-word, pixel-pixel, and pixel-word.
The Fully Convolutional Network (FCN)-like framework~\cite{long2015fully,ding2020semantic} is commonly used in existing referring segmentation methods~\cite{hu2016segmentation,margffoy2018dynamic}. They usually perform convolution operations on the fused, \eg, concatenated or multiplied, vision-language features to predict the segmentation mask for the target object. However, the long-range dependency modeling is intractable by regular convolution operation as its large receptive field is achieved by stacking many small-kernel convolutions. 
This oblique process makes the information interaction between long-distance pixels/words inefficient~\cite{wang2018non}, thus is undesirable for the referring segmentation model to understand the global context expressed by the input image and language~\cite{ye2019cross}. In recent years, attention mechanism has gained considerable popularity in the computer vision community thanks to its advantage in building direct interaction among all elements, which greatly helps the model in capturing global semantic information.
There have been some previous referring segmentation works that use attention to alleviate the long-range dependency issues, \eg, ~\cite{ye2019cross,hu2020bi,shi2018key}. 
However, most of them rely on FCN-like pipelines~\cite{ding2018context,ding2019boundary} and only use the attention mechanism as auxiliary modules, which limits their ability to model the global context.
In this work, we reformulate the referring segmentation problem as a direct attention problem and re-construct the current FCN-like framework using Transformer~\cite{vaswani2017attention}. We generate a set of query vectors from language features using vision-guided attention, and use these vectors to ``query'' the given image and predict the segmentation mask from the query responses, as shown in \figurename~\ref{fig:fig1}. This attention-based framework enables us to implement global operation among multi-modal features in each computation stage and enhances the network's ability to capture the global context of both vision and language information.

Secondly, in order to handle the randomness caused by the various objects/images and the unrestricted language expressions, we propose to understand the input language expression in different ways incorporating vision features. In many existing referring segmentation methods, such as~\cite{luo2020multi, yu2018mattnet}, the language self-attention is often used to extract the most informative part and emphasized word(s) in the language expression.
However, for these methods, their language understanding is derived solely from the language expression itself without interacting with the vision information. As a sequence, they cannot distinguish which emphasis is more suitable and effective that can fit a particular image better. Hence, their detected emphases might be inaccurate or inefficient. On the other hand, in most existing vision-transformer works~\cite{carion2020end}, the queries of the transformer decoder are a set of fixed and learned vectors, each of which predicts an object. However, experiments show that each query vector has its own operating modes, and is specifically targeted at certain kinds of objects~\cite{carion2020end}, \eg, specifically targeted at objects of a certain type or located in a certain area.
The fixed queries in these works implicitly assume that the objects in the input image are distributed under some certain statistical rules, which does not well consider the randomness and huge diversity of the referring segmentation, especially the randomness brought by unconstrained language expressions. Besides, the learnable queries are designed for detecting all the objects in the whole image instead of focusing on the target object indicated by the language expression, thus cannot efficiently extract informative representation that contains the clues to the target object.
To address these issues, we propose to generate input-specific queries that could focus on the clues related to the referred target object. We herein propose a Query Generation Module (QGM), which dynamically produces multiple query vectors based on the input language expression and the vision features. Each query vector represents a specific comprehension of the language expression and queries the vision features with different emphases. As shown in \figurename~\ref{fig:fig1}, three queries focus on different information, respectively. 
These generated query vectors produce a set of corresponding masks in the transformer decoder though we only need one mask selected from them. Besides, we also hope to choose a more reasonable and better comprehension way from these query vectors. Therefore, we further propose a Query Balance Module (QBM), which assigns each query vector a confidence measure to control its impact on mask decoding, and then adaptively selects the output features of these queries to better generate the final mask. The proposed QGM dynamically produces input-specific queries that focus on different informative clues related to the target object, while the proposed QBM selectively fuses the corresponding responses by these queries. These two modules work together to prominently improve the diverse ways to understand the image and query language and enhance the network's robustness towards highly random inputs.

Thirdly, we introduce masked contrastive representation learning to further enhance the model's generalization ability and robustness to unconstrained language expressions. With the proposed Query Generation Module and Query Balance Module, we provide different understandings of a given expression, which can be viewed as a kind of intral-sample learning. Here we further consider inter-sample learning to explicitly endow the model with knowledge of different language expressions to one object. For the same target object, there are multiple ways to describe it. However, the final representations that predict the target mask should be the same. In other words, the output features of Query Balance Module by different expressions for the same object should be the same.
To this end, we utilize contrastive learning to narrow down the features of different expressions for a same target object, while distinguishing the features of different objects. What's more, we observe that the model tends to overly rely on specific words that provide the most discriminative clues or frequently occur in training samples, while ignoring other complementary information. The excessive reliance on specific words will damage the model's generalization ability, for instance, the model may not well understand testing expressions that do not contain common discriminative clues in the training samples. To address this issue, we introduce masked language expressions in contrastive representation learning, which randomly erases some specific words from the original language expression. The masked language expression and the original expression refer to the same target object, they are considered as a positive pair in the contrastive representation learning to be close to each other and reach the same representation. The masked contrastive representation learning significantly enhances the model's ability in dealing with diverse language expressions in the wild.

The proposed approach builds deep interactions between language and vision information at different levels, which greatly enhances the utilization and fusion of multi-modal features. Besides, the proposed network is lightweight and its parameter scale is roughly equivalent to just seven convolution layers. In summary, our main contributions are listed as follows:
\begin{itemize}
\setlength\itemsep{0.3em}
   \item We design a Vision-Language Transformer~(VLT) framework to facilitate deep interactions among multi-modal information and enhance the holistic understanding to vision-language features.
   \item We propose a Query Generation Module (QGM) that dynamically produces multiple input-specific queries representing different comprehensions of the language, and a Query Balance Module (QBM) to selectively fuse the corresponding responses by these queries. 
   \item We introduce a masked contrastive representation learning to enhance the model's generalization ability and robustness to deal with the unconstrained language expressions by learning inter-sample relationships.
   \item The proposed approach is lightweight and achieves new state-of-the-art performance consistently on three referring image segmentation datasets, RefCOCO, RefCOCO+, G-Ref, and two referring video object segmentation datasets, YouTube-RVOS and Ref-DAVIS17.
\end{itemize}

%------------------------------------------------------------------------
\section{Related Works}
In this section, we discuss works that are closely related to the proposed approach, including referring segmentation, referring comprehension, and transformer. %\xlnote{spatial/temporal propagation network}
%-------------------------------------------------------------------------
\subsection{Referring Segmentation}

Referring segmentation is one of the most fundamental while challenging multi-modal tasks, involving both language and vision information. Given a natural language expression describing the properties of the target object in the given image, the goal of the referring segmentation is to ground the target object referred by the language and generate a corresponding segmentation mask. Inspired by the task of referring comprehension~\cite{wang2019neighbourhood,liu2019learning,yang2019fast,zhuang2018parallel,yang2020improving,liao2020real}, referring segmentation is introduced by Hu \etal in~\cite{hu2016segmentation}. \cite{hu2016segmentation} concatenates the linguistic features extracted by Long Short-Term Memory (LSTM) networks and the visual features extracted by Convolutional Neural Networks (CNN). Then, the fused vision-language features is inputted to a fully convolutional network (FCN)~\cite{long2015fully} to generate the target segmentation mask. In~\cite{liu2017recurrent}, in order to better utilize the information of each word in the language expression, Liu \etal propose a multimodal LSTM (mLSTM), which models each word in every recurrent stage to fuse the word feature with vision features. Li \etal~\cite{li2018referring} utilize features from different levels in the backbone progressively, which further improves the performance. To better utilize the language information, Edgar \etal~\cite{margffoy2018dynamic} propose a method that uses the feature of each word in the language expression when extracting language features, not just the final state of the RNN. Chen \etal~\cite{Chen_lang2seg_2019} employ a caption generation network to produce a caption sentence that describes the target object, and enforce the caption to be consistent with the input expression. In~\cite{luo2020multi}, Luo \etal propose a multi-task framework to jointly learn referring expression comprehension and segmentation. They build a network that contains a referring expression comprehension branch and a referring expression segmentation branch, each of which can reinforce the other during training. Jing~\etal~\cite{jing2021locate} decouple the referring segmentation to localization and segmentation and propose a Locate-Then-Segment (LTS) scheme to locate the target object first and then generate a fine-grained segmentation mask. Feng \etal~\cite{feng2021encoder} propose to utilize the language feature earlier in the encoder stage. Hui \etal~\cite{hui2020linguistic} introduce a linguistic structure-guided context modeling to analyze the linguistic structure for better language understanding. Yang \etal~\cite{yang2021bottom} propose a Bottom-Up Shift (BUS) to progressively locates the target object with hierarchical reasoning the given expression. 

With the introduction of attention-based methods~\cite{wang2018non,vaswani2017attention}, researchers have found that the attention mechanism is suitable for the formulation of referring segmentation. For example, Ye \etal propose the Cross-Modal Self-Attention (CMSA) model~\cite{ye2019cross} to dynamically find the most important words in the language sentence and the informative image region. Hu \etal~\cite{hu2020bi} propose a bi-directional attention module to further utilize the features of words. Most of these works are built on FCN-like networks~\cite{ding2021interaction,ding2019semantic} and only use the attention as auxiliary modules. Our concurrent work MDETR~\cite{kamath2021mdetr} employs DETR~\cite{carion2020end} to build an end-to-end modulated detector and reason jointly over language and image. After the proposed VLT~\cite{ding2021vision}, transformer-based referring segmentation architectures receive more attention~\cite{MaIL, yang2021lavt, wang2022cris, jain2021comprehensive, kim2022restr}. MaIL~\cite{MaIL} follows the transformer architecture ViLT~\cite{kim2021vilt} and utilizes instance mask predicted by Mask R-CNN~\cite{he2017mask} as additional input. Yang \etal~\cite{yang2021lavt} propose Language-Aware Vision Transformer (LAVT) to conduct multi-modal fusion at intermediate levels of the network. CRIS~\cite{wang2022cris} employs CLIP~\cite{radford2021learning} pretrained on 400M image text pairs and transfers CLIP from text-to-image matching to text-to-pixel matching.

\begin{figure*}[t]
   \begin{center}
      \includegraphics[width=1\linewidth]{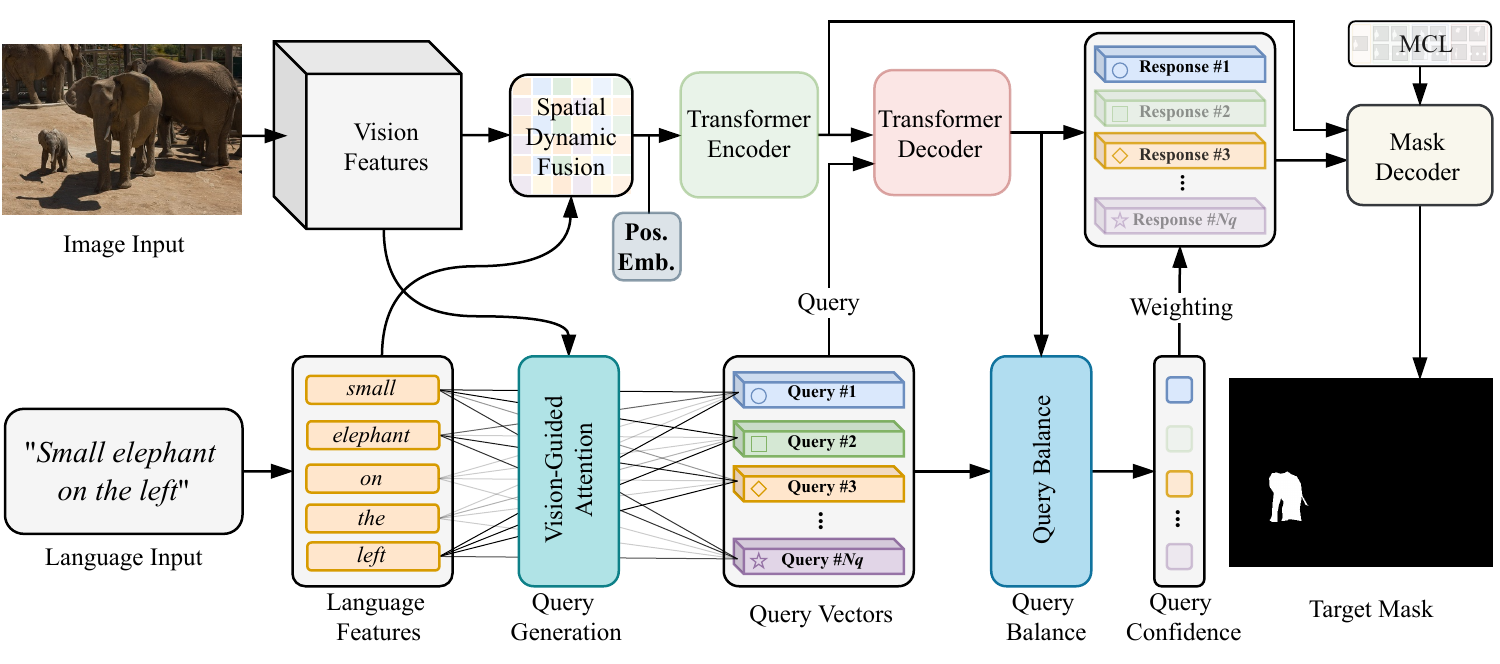}
   \end{center}
    \vspace{-0.16in}
   \caption{The overview architecture of the proposed Vision-Language Transformer (VLT). Firstly, the given image and language expression are projected into visual and linguistic feature spaces, respectively. A Spatial Dynamic Fusion module is then employed to fuse vision and language features, generating multi-modal feature inputted to the transformer encoder. The proposed Query Generation Module generates a set of input-specific queries according to the vision and language features. These input-specific queries are sent to the decoder, producing corresponding query responses. These resulting responses are selected by the Query Balance Module and then decoded to output the target mask by a Mask Decoder. ``Pos. Emb.'': Positional Embeddings. ``MCL'': Masked Contrastive Learning.}
   \label{fig:network}
\end{figure*}

In this work we employ a fully attention-based architecture and propose a Vision-Language Transformer (VLT)~\cite{ding2021vision} to model the long-range dependencies in the image, as shown in \figurename~\ref{fig:network}. We further propose to generate input-conditional queries for the decoder of transformer to better understand the unconstrained language expressions from different aspects.

\subsection{Referring Comprehension}

Referring comprehension is a highly relevant task to referring segmentation. Referring comprehension also takes an image and a language expression as inputs and identifies the target object referred by the language expression. However, while referring segmentation aims to output a segmentation mask for the target object, the referring comprehension outputs a grounding box. Unlike the FCN-like pipeline of referring segmentation, most earlier referring comprehension works are based on the multi-stage pipeline\cite{liu2019learning, yang2019fast,zhuang2018parallel,hu2017modeling,zhang2018grounding,hong2019learning}. In these works, often an out-of-the-box instance segmentation network, \eg, Mask R-CNN~\cite{he2017mask}, is first applied to the image and generates a set of instance proposals, regardless of the language input. Next, the candidate proposals are compared with the language expression, to find the best match. For example, Yu \etal~\cite{yu2018mattnet} propose a two-stage method that first extracts all instances in the image using Mask R-CNN~\cite{he2017mask}, then employs a modular network to match and select the target object from all the instances detected by Mask R-CNN. In recent years, one-stage methods \cite{chen2018real, liao2020real, deng2021transvg} have also been increasingly adopted in the referring comprehension area, \eg, Sadhu \etal propose a ``Zero-Shot Grounding'' network for referring comprehension \cite{sadhu2019zero}, and Yang \etal design a recursive sub-query construction framework to gradually reason between the image and query language \cite{yang2020improving}.

\subsection{Transformer}
Transformer is first proposed by Vaswani \etal in~\cite{vaswani2017attention} for machine translation, it is a sequence-to-sequence deep network architecture with the attention mechanism. Recently, transformer attracts lots of attention in Natural Language Processing (NLP) and achieves great success on many NLP tasks, \eg, machine translation~\cite{vaswani2017attention}, question answering~\cite{devlin2018bert}, and language modeling~\cite{krause2019dynamic}. Besides the NLP, transformer has also been employed to many computer vision tasks~\cite{cai2022degradation,lin2022flow} and has achieved promising results on various vision tasks such as object detection~\cite{carion2020end}, image recognition~\cite{dosovitskiy2020image}, semantic segmentation~\cite{zheng2020rethinking, xie2021segformer,liang2022expediting}, and human-object interaction~\cite{wangsuchen_iccv2021, wang2022learning}.

In the vision-language field, transformer architectures have achieved great success in many tasks, \eg, vision-and-language pre-training~\cite{Vilbert, kim2021vilt,Vinvl}, image generation~\cite{DALL}, visual question answering~\cite{huang2020pixel}, open-vocabulary detection~\cite{OVR-CNN}, image retrieval~\cite{FashionVLP}, vision-and-language navigation~\cite{HAMT}, \etc Lu \etal~\cite{Vilbert} design a co-attention mechanism to incorporate language-attended vision features into language features. Kim \etal~\cite{kim2021vilt} propose to deal with the two modalities in a single unified transformer architecture. Huang \etal~\cite{huang2020pixel} propose a Pixel-BERT to align visual features with textual features by jointly learning visual and textual embedding in a unified transformer way. Based on the Pixel-BERT, Zareian \etal~\cite{OVR-CNN} design a vision-to-language projection to process visual features before transformer. Radford~\etal~\cite{radford2021learning} propose a Contrastive Language-Image Pre-training (CLIP) scheme to jointly train image language encoders. Wang \etal~\cite{wang2022cris} apply CLIP to referring image segmentation by text-to-pixel alignment. Lei~\etal~\cite{lei2021less} introduce a C{\footnotesize LIP}BERT to text-to-video retrieval and video question answering. Hu~\etal~\cite{Hu_2021_ICCV_UniT} propose a Unified Transformer (UniT) model to learn multiple vision-language tasks with a unified multi-modal architecture. Different from previous works that use a small fixed number of learned position embeddings as object queries, we propose to dynamically generate input-specific queries representing different comprehensions of language and selectively fuse the corresponding responses by these input-specific queries. With the input-specific queries, the proposed VLT better captures the informative clues hidden in the language expressions and address the high diversity in referring segmentation.

%-------------------------------------------------------------------------

\section{Methodology}
The overall architecture of the proposed Vision-Language Transformer (VLT) is shown in \figurename~\ref{fig:network}. The network takes a language expression and an image as inputs. First, the input image and language expression are projected into the linguistic and visual feature spaces, respectively. Then, vision and language features are inputted to the proposed Query Generation Module (QGM) to generate a set of input-specific query vectors, which represent different understandings of the language expression under the guidance of visual clues. Meantime, vision and language features are fused to multi-modal feature by the proposed Spatial Dynamic Fusion (SDF), and the multi-modal feature is sent to the transformer encoder to produce a group of memory features. The query vectors in Q generated by our proposed QGM are employed to ``query'' K and V derived from memory features in transformer decoder, \ie, $\text{Attention}(Q, K, V ) = \text{softmax}(\frac{QK^T}{\sqrt{d_k}})V$, where $d_k$ is the dimensionality of $K$. The resulting responses from transformer decoder are then selected by a Query Balance Module (QBM) with different confidence weights. Finally, the mask decoder takes the weighted responses from QBM and the output feature from transformer encoder as inputs and outputs a mask for the target object. Masked Contrastive Learning (MCL) is used to supervise the features in Mask Decoder to narrow down the features of different expressions for the same target object while distinguishing the features of different objects. Positional embeddings are used to supplement the pixel position information in the permutation-invariant transformer architecture.

\begin{figure}[t]
   \begin{center}
     \includegraphics[width=0.996\linewidth]{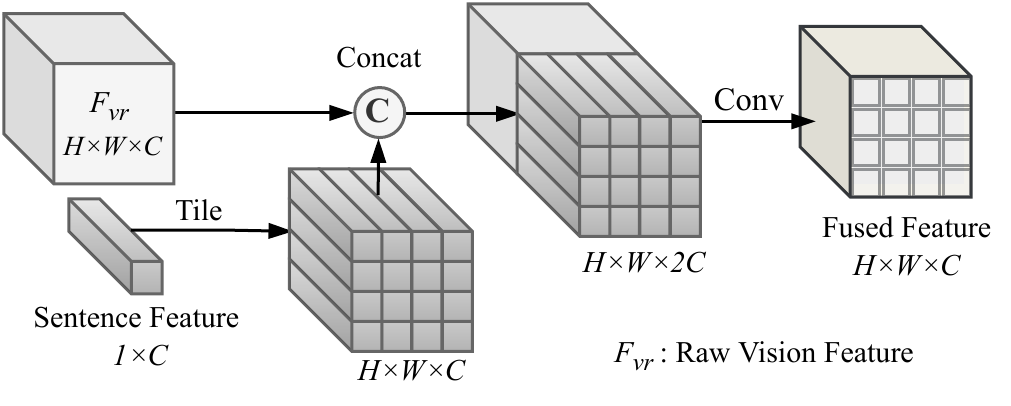}
   \end{center}
  \vspace{-0.16in}
   \caption{Tile-and-concatenate fusion. The language feature is identically copied to every position across the $H\times W$ map.}
   \label{fig:tile-and-concatenate}
\end{figure}

\begin{figure*}[htbp]
   \begin{center}
     \includegraphics[width=\linewidth]{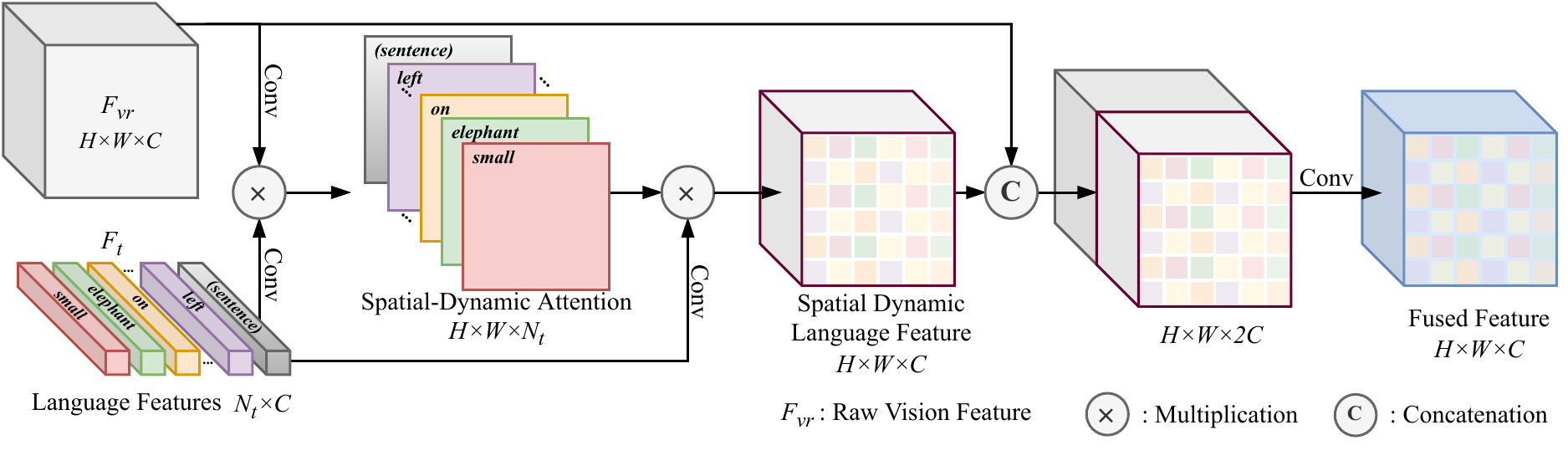}
   \end{center}
  \vspace{-0.16in}
   \caption{An illustration of the proposed Spatial-Dynamic Fusion (SDF). Different from the conventional ``tile-and-concatenate'' fusion, the proposed SDF finds a word attention set and derives a tailored language feature vector for each pixel in the image feature.}
   \label{fig:sdf}
\end{figure*}

\subsection{Spatial-Dynamic Multi-Modal Fusion}\label{sec:sdf}

After backbone features for language and image are extracted, the first step is to preliminarily fuse them together and generate a multi-modal feature. 
For referring segmentation, effective multi-modal information fusion is critical and challenging because the unconstrained expression of natural language and the diversity of objects in scene images bring huge uncertainty to the understanding and fusion of multi-modal features.
However, existing approaches conduct multi-modal feature fusion simply by either concatenation~\cite{hu2016segmentation,liu2017recurrent,ye2019cross,hu2020bi} or point-wise multiplication~\cite{luo2020multi,ding2021vision,jing2021locate} of vision feature and language feature. The language feature, which is a 1D vector, is usually tiled to every position of the vision feature~\cite{hu2016segmentation,ye2019cross}, as shown in \figurename~\ref{fig:tile-and-concatenate}. Under the ``tile-and-concatenate'' operation, the language feature is identically copied to every position across the $H\times W$ map.

Although such kinds of fusion techniques are simple and have achieved reasonable performance, there are a few drawbacks. Firstly, the features of individual single words are not fully utilized in this step. Secondly, the tiled language feature will be identical for all pixels across the image feature, which weakens the location information carried by the correlation between the language information and the visual information. Due to the diversity of objects in the input image, an image usually contains diverse information that can be very complex, where different regions may contain different semantic information. Meanwhile, the language expression can be interpreted with different emphases from different perspectives. We here emphasize the differences among pixels/objects, \ie, the vision information across the image varies from place to place. Therefore, the informative words in a given sentence are different from pixel to pixel.
The way of tiling ignores such differences and simply assigns the same language feature vector to every pixel, resulting in some confusion. It is better to make tailored feature fusion specifically for each individual pixel. In this work, we propose a Spatial-Dynamic Fusion (SDF) module, which produces different language feature vectors for different positions of the image feature according to the interaction between language information and corresponding pixel information. Each position selects its interested words and pays more attention to these words during multi-modal fusion.

An illustration of the proposed Spatial-Dynamic Fusion (SDF) module is shown in \figurename~\ref{fig:sdf}. The proposed SDF module takes language features $F_t$, including features of each word and the whole sentence, and image feature $F_{vr}$ as inputs. We first use language feature and vision feature to generate the Spatial-Dynamic Attention matrix by:
\begin{equation}
   A_{\mathrm{sd}} = \mathrm{softmax}(\frac{1}{\sqrt{C}}\mathrm{Conv}(F_{vr})\mathrm{Conv}(F_t)^T),
\end{equation}
where $C$ is feature channel number and $\frac{1}{\sqrt{C}}$ is the scaling factor. $A_{\mathrm{sd}}$ is with the shape of $H~\!\!\times~\!\!W\times~\!\!N_t$, where $H$ and $W$ are height and width respectively, and $N_t$ is the number of language feature vectors in $F_t$. Softmax normalization is applied along the $N_t$ axis of the attention matrix $A_{\mathrm{sd}}$. Each position of the spatial-dynamic attention $A_{\mathrm{sd}}$ is a weighting vector that indicates different importance of the $N_t$ language features at this position. Therefore, a spatial dynamic language feature $F_{\mathrm{sdl}}$ is generated by:
\begin{equation}
      F_{\mathrm{sdl}} = A_{\mathrm{sd}}\mathrm{Conv}(F_t),
\end{equation}
where $F_{\mathrm{sdl}}$ is in the shape of $H~\!\!\times~\!\!W\times~\!\!C$, each vector of $F_{sdl}$ across $H~\!\!\times~\!\!W$ is the language feature vector weighted by its correlation to the image context at a pixel position. The fused multi-modal feature $F_{\mathrm{fused}}$ is generated by:
\begin{equation}
      F_{\mathrm{fused}} = \mathrm{Conv}(F_{\mathrm{sdl}}\copyright F_{vr}),
\end{equation}
where $\copyright$ denotes concatenation.
Following previous transformer works~\cite{carion2020end,bello2019attention}, we employ fixed sine spatial positional embeddings to supplement the pixel position information in the permutation-invariant transformer.
The fused multi-modal feature and the positional embeddings are inputted to the transformer encoder (see \figurename~\ref{fig:network}). 

\begin{figure}[t]
   \centering
   \footnotesize{\hspace{4em} \textbf{Input: \textit{``The large balloon on the left"}} \hspace{4em} } \\
   \vspace{0.35em}
   \hfill
   \begin{subfigure}[b]{0.42\linewidth}
       \centering
       \includegraphics[width=0.9\textwidth]{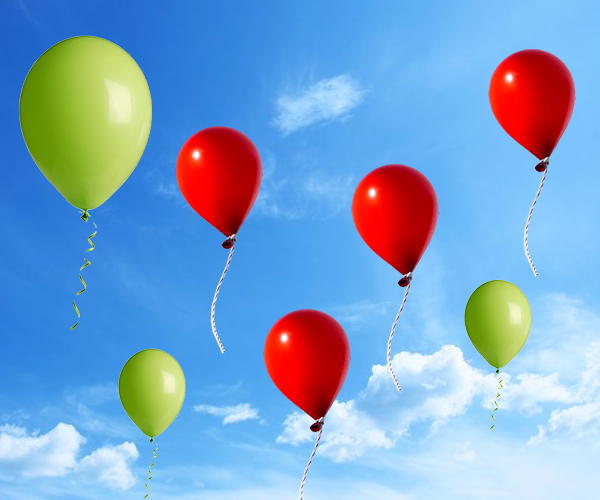}
       \caption{}
   \end{subfigure}
   \hfill
   \begin{subfigure}[b]{0.42\linewidth}
       \centering
       \includegraphics[width=0.9\textwidth]{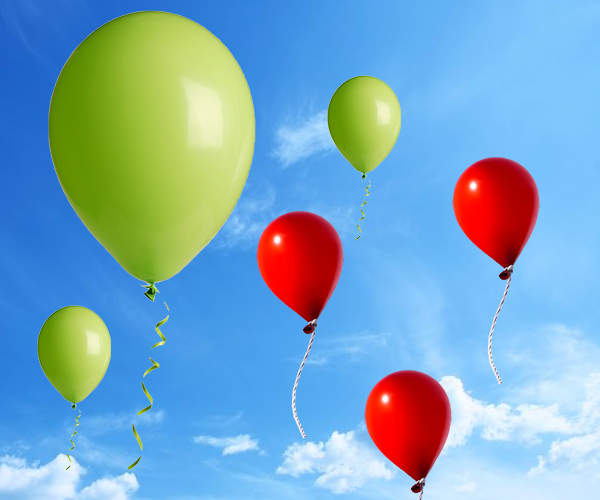}
       \caption{}
   \end{subfigure}
   \hfill
  \vspace{-0.06in}
   \caption{An example of one sentence with different emphasis. For different images, the informative degree of words ``large'' and ``left'' are different.}
   \label{fig:q_eg}
\end{figure}
%-------------------------------------------------------------------------
\subsection{Query Generation Module}\label{sec:QGM}

In most existing Vision Transformer works, \eg,~\cite{carion2020end,zheng2020rethinking, xie2021segformer,cheng2021maskformer}, queries for the transformer decoder are usually a set of fixed learned vectors, each of which is used to predict one object and has its own operating mode, \eg, specifying objects of a certain kind or located in a certain region. These works with fixed queries have an implicit assumption that objects in the input image are distributed under some statistic rules. However, such an assumption does not consider the huge diversity of the referring segmentation. Besides, the learnable queries are designed for detecting all objects in the whole image instead of focusing on the target object indicated by the language expression, thus cannot effectively extract informative representation that contains clues of the target object.

\begin{figure}[t]
   \begin{center}
      \includegraphics[width=0.936\linewidth]{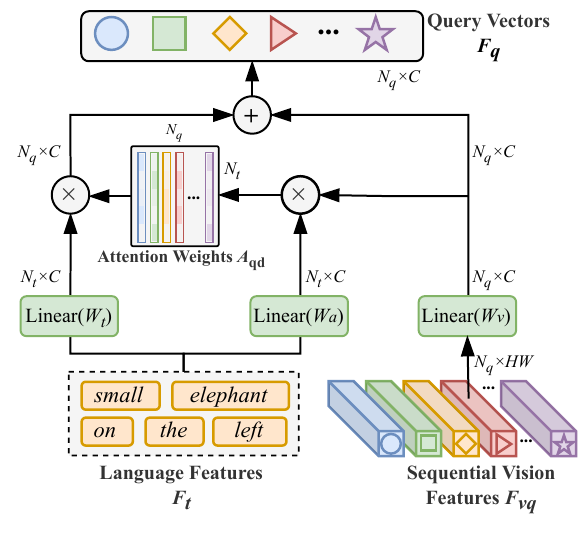}
   \end{center}
    \vspace{-0.1in}
   \caption{Query Generation Module (QGM). The QGM takes sequential vision feature $F_{vq}$ and language features $F_t$ as inputs and generates a group of input-specific query vectors $F_q$, which are then sent to the transformer decoder of our VLT.}
   \vspace{-0.1in}
   \label{fig:query_gen}
\end{figure}
For referring segmentation, the target object described by the given expression can be any part of the image. Because both the input image and language expression are unconstrained, the {stochasticity} of the target object's properties is significantly high. Therefore, fixed query vectors, like in most existing ViT works, cannot well represent the properties of the target object. Instead, the properties of the target object are hidden in the input language expression, \eg, keywords like {``blue/yellow''}, {``small/large''}, {``right/left''}, \etc To capture the informative clues and address the high {stochasticity} in referring segmentation, we propose a Query Generation Module (QGM) to adaptively generate the input-specified query vectors online according to the given image and language expression. Also, it is well known that for a language expression, the importance of different words is different. Some existing works address this issue by measuring the importance of each word. For example,~\cite{luo2020multi} gives each word a weight and~\cite{yu2018mattnet,huang2020referring} defines a set of groups, \eg, location, attribute, entity, and finds the degree of each word belonging to different groups. Most works derive the weights by the language self-attention, which does not utilize the information in the image and only outputs one set of weights. But in practice, a same sentence may have different understanding perspectives and emphasis, and the most suitable and effective emphasis can only be known with the help of the image. We give an intuitive example in \figurename~\ref{fig:q_eg}. For the same input sentence \textit{``The large balloon on the left''}, the word \textit{``left''} is more informative for the first image while the word \textit{``large''} is more useful for the second image. In this case, language self-attention cannot differentiate the importance between \textit{``large''} and \textit{``left''}, making the attention process less effective. In order to let the network learn different aspects of information and enhance the robustness of the queries, we generate multiple queries with the help of visual information, though there is only one target instance. Each query represents a specific comprehension of the given language expression with different emphasized word(s).

\begin{figure}[t]
   \begin{center}
      \includegraphics[width=1\linewidth]{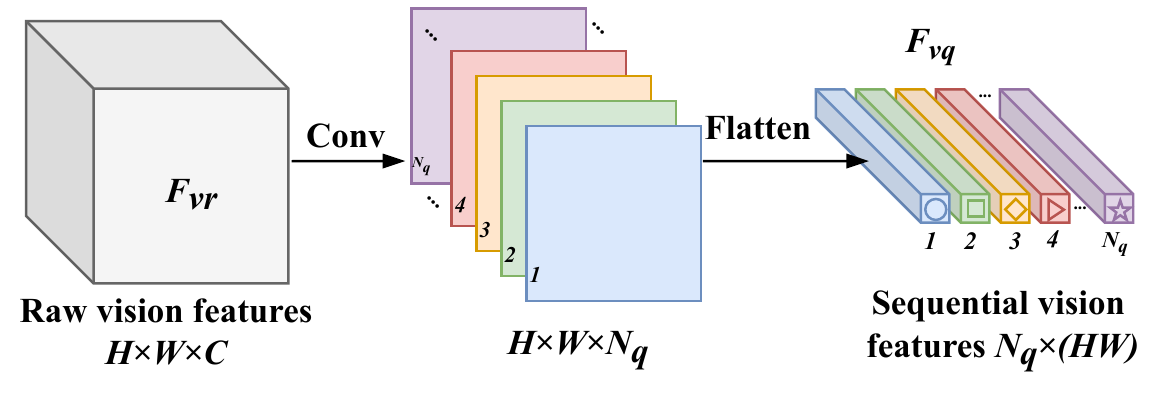}
   \end{center}
  \vspace{-0.1in}
   \caption{The preparation process of the sequential vision features for our Query Generation Module.}
   %\vspace{-0.1in}
   \label{fig:fv_prep}
\end{figure}

\begin{figure*}[t]
   \begin{center}
      \includegraphics[width=0.96\linewidth]{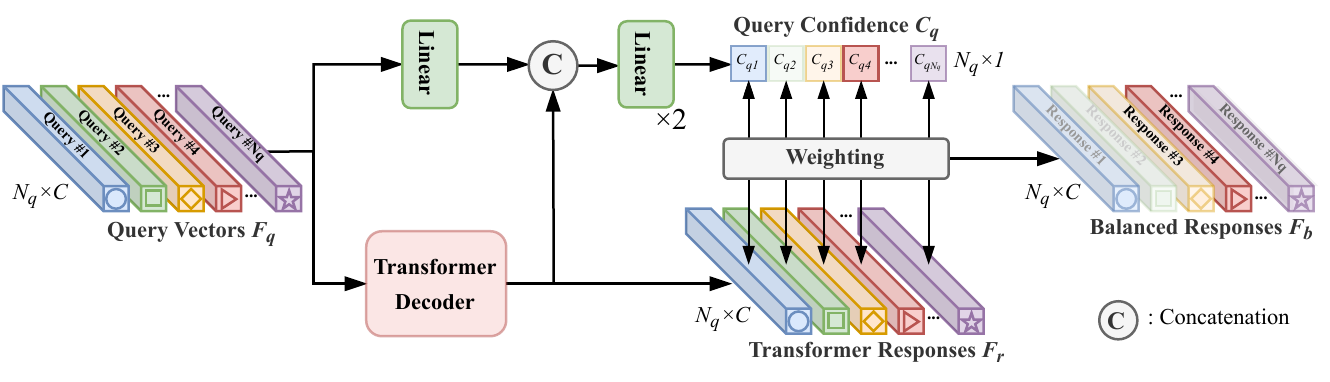}
   \end{center}
    \vspace{-0.1in}
   \caption{Query Balance Module (QBM). For each query vector, a confidence measure parameter $C_q$ is computed to reflect how much it fits the prediction and the context of the image. The transformer responses $F_r$ is weighted by the corresponding confidences $C_q$ to control the influence of each query vector, generating balanced responses $F_b$.}
   \label{fig:query_weight}
\end{figure*}
The architecture of the Query Generation Module is shown in \figurename~\ref{fig:query_gen}. It takes language feature $F_t\in \mathbb{R}^{N_t\times C}$ and raw vision feature $F_{vr}\in \mathbb{R}^{H\times W\times C}$ as inputs. In $F_{t}$, the $i$-th vector is the feature vector of the word $w_i$, which is the $i$-th word in the input language expression. $N_t$ denotes the sentence length and is fixed over all inputs by zero-padding. This module aims to output $N_q$ query vectors, each of which is a language feature with different attention weights guided by the vision information. Specifically, the vision features are firstly prepared as shown in \figurename~\ref{fig:fv_prep}. We reduce the feature channel dimension size of the raw vision feature $F_{vr}$ to query number $N_q$ by three convolution layers, resulting in $N_q$ feature maps. Each of them will participate in the generation of one query vector. The feature maps are then flattened in the spatial domain, forming a feature matrix $F_{vq}$ of size $N_q\times (HW)$, \ie,
\begin{equation}
   F_{vq}=\text{Flatten}(\text{Conv}(F_{vr}))^T.
\end{equation}

Next, we comprehend the language expression from multiple aspects incorporating the image, forming $N_q$ queries from language. We derive the attention weights for language features $F_t$ by incorporating the vision features $F_{vq}$,
\begin{equation}
   A_\mathrm{qd}=\mathrm{softmax}(\frac{1}{\sqrt{C}}{\sigma(F_{vq}W_v)\sigma(F_{t}W_a)^T}),
\label{eq:5}
\end{equation}
where $A_\mathrm{qd}\!\in\!\mathbb{R}^{N_q\times N_t}$ is query-dynamic attention matrix, containing $N_q$ different attention vectors to $F_t$. $W_v\in \mathbb{R}^{(HW)\times C}$ and $W_a\in \mathbb{R}^{C\times C}$ are learnable parameters, $\sigma$ is activation function Rectified Linear Unit (ReLU). The softmax function is applied across all words for each query as normalization. In attention matrix $A_\mathrm{qd}$, each of the $N_q$ vectors consists of a set of attention weights for different words. Different queries attend to different parts of the language expression. Thus, $N_q$ query vectors focus on different emphasis or different comprehension ways of the language expression. Notably, after this step, for longer sentences, we randomly mask one of the most important words to enhance the generalization ability of the network. The details are shown in Sec. \ref{sec:misl}.

Next, the derived attention weights are applied to the language features:
\begin{equation}
   F_{q}=A_\mathrm{qd}\sigma(F_{t}W_t) +\sigma(F_{vq}W_v),
\label{Eq:Fq}
\end{equation}
where $W_t\in \mathbb{R}^{C\times C}$ and $W_v$ are learnable parameters, $F_q\in \mathbb{R}^{N_q\times C}$ contains $N_q$ query vectors $\{F_{q1},...,F_{qn},...F_{qN_q}\}$. We add a residual connection from vision feature $F_{vq}$ to enrich the information in query vectors. Each $F_{qn}$ is an attended language feature vector guided by vision information and serves as one query vector to the transformer decoder.

%-------------------------------------------------------------------------
\subsection{Query Balance Module}\label{sec:QBM}
We get $N_q$ different query vectors from the proposed Query Generation Module. Each query represents a specific comprehension of the input language expression under the interactive guidance of the input image information. As we discussed before, both the input image and language expression are of high {arbitrarines}. Thus, it is desired to adaptively select the better comprehension ways and let the network focus on the more reasonable and suitable comprehension ways.
On the other hand, as the independence of each query vector is kept in the transformer decoder~\cite{carion2020end} but we only need one mask output, it is desired to balance the influence of different queries on the final output. Therefore, we propose a Query Balance Module (QBM) to dynamically assign each query vector a confidence measure that reflects how much it fits the prediction and the context of the image.

The architecture of the proposed QBM is shown in \figurename~\ref{fig:query_weight}. Specifically, the inputs of Query Balance Module are the query vectors $F_{q}$ from the Query Generation Module and its corresponding responses from the transformer decoder, $F_{r}$, which is of the same size as $F_{q}$. In the Query Balance Module, the query vectors after going through a linear layer and their corresponding responses are first concatenated together. The linear layers are employed to derive confidence levels according to the query vectors $F_{q}$ and their corresponding responses $F_{r}$. Then, a set of query confidence levels $C_q$, in the shape of $N_q\times 1$, are generated by two consecutive linear layers. Sigmoid, $S(x)=\frac{1}{1+e^{-x}}$, is employed after the the last linear layer as an activation function to control the output range. Let $F_{rn}$ and $C_{qn}$ denote the corresponding response and query confidence to the $n$-th query $F_{qn}$, respectively. Each scalar $C_{qn}$ shows how much the query $F_{qn}$ fits the context of its prediction, and controls the influence of its response $F_{rn}$ to the mask decoding. Each response $F_{rn}$ is multiplied with the corresponding query confidence $C_{qn}$, \ie, $F_{bn}=F_{rn}C_{qn}$. The balanced responses $F_{b}=\{F_{b1},...,F_{bn},...,F_{bN_q}\}$ are sent for mask decoding. The proposed QGM dynamically produces input-specific queries that focus on different informative clues related to the target object, while the proposed QBM selectively fuses the corresponding responses to these queries. These two modules work together to prominently boost the diversity to understand the image and query language, and enhance the model's robustness towards highly stochastic inputs.

\begin{figure}[t]
   \begin{center}
      \includegraphics[width=0.916\linewidth]{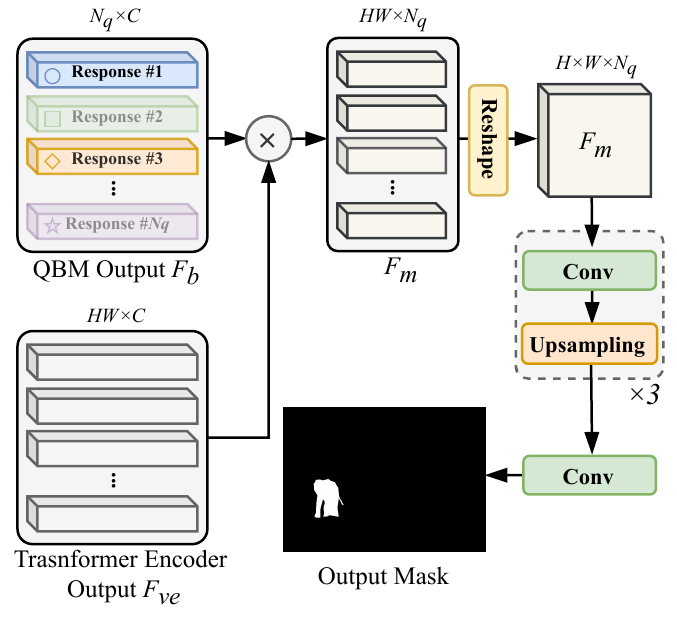}
   \end{center}
  \vspace{-0.1in}
   \caption{The Mask Decoder takes the outputs of Query Balance Module (QBM) $F_b$ and Transformer Encoder $F_{ve}$ to generate the output mask.}
   \label{fig:mask_decoder}
\end{figure}

\begin{figure*}[htbp]
   \begin{center}
      \includegraphics[width=0.9\linewidth]{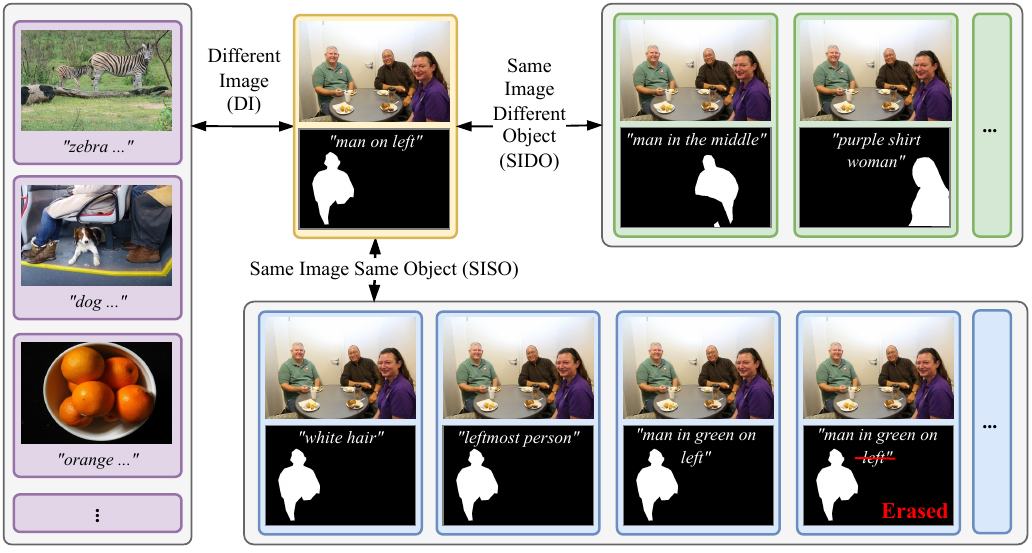}
   \end{center}
  \vspace{-0.16in}
   \caption{Different kinds of inter-sample relationships. SISO: Same Image, Same Object (but different expressions). SIDO: Same Image, Different Object. DI: Different Image. We erase some common word(s) in the long sentences and add such samples into SISO.}
   \label{fig:isl}
\end{figure*}

%-------------------------------------------------------------------------
\subsection{Mask Decoder}\label{sec:maskdecoder}
The output of the Query Balance Module $F_b$ with the size of $N_q\times C$ is sent to the mask decoder, as shown in \figurename~\ref{fig:mask_decoder}. In the mask decoder module, $F_b$ is utilized as a set of mask generation kernel to process the vision-dominated feature $F_{ve}$ from the transformer encoder, to produce mask feature $F_{m}$, \ie,
\begin{equation}
    F_{m} = F_{ve}F_b^T,\label{eq:maskdecoder}
\end{equation}
where $F_{ve}$ is with size of $HW\times C$ so that $F_{m}$ has size of $HW\times N_q$. Then we reshape $F_{m}$ to $H\times W\times N_{q}$ for the final mask generation. We use three stacked $3\times 3$ convolution layers for decoding followed by one $1\times1$ convolution layer for outputting the final predicted segmentation mask. To control the output size and generate a higher-resolution mask, upsampling layers are placed after each of the three $3\times 3$ convolution layers. To better demonstrate the effectiveness of the proposed transformer module, the Mask Decoding Module in our implementation does not utilize any CNN features. We employ the Binary Cross-Entropy loss on the predicted masks to supervise the network training.

\begin{figure}[t]
   \begin{center}
      \includegraphics[width=1\linewidth]{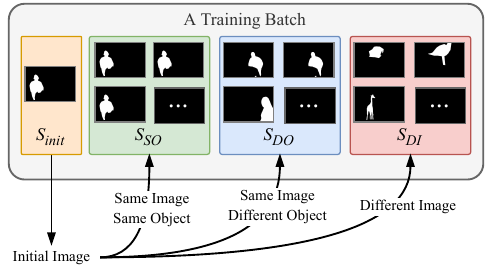}
   \end{center}
   \vspace{-0.1in}
   \caption{One training batch in inter-sample learning.}
   %\vspace{-0.1in}
   \label{fig:isl_batch}
\end{figure}
%-------------------------------------------------------------------------
\subsection{Masked Contrastive Learning}\label{sec:misl}
Here we further consider inter-sample learning to explicitly endow the model with the knowledge of different language expressions to one object. The given expression in natural language is unconstrained. There are multiple ways to describe the same target object, which brings challenges in understanding these expressions. Given an image $I$ that contains $N_O$ objects $\{O_1,...,O_i,...,O_{N_O}\}$, every object $O_i$ in $I$ can be referred by ${N_E}$ different expressions $\{E_i^1,...,E_i^j,...,E_{i}^{N_E}\}$. A sample $S(I,O_i,E_i^j)$ of referring segmentation defines a mapping from an expression to the target object: $\{E_i^j\rightarrow O_i| I\}$. The mappings between $E_i$ and $O_i$ are in general many-to-one. An object in the image can be described by many different language expressions, but one language expression should unambiguously point to one and only one instance. Thus, no matter what kind of expressions are given to the target object, the final mask is the same, \ie, the feature $F_m$ in Eq.~(\ref{eq:maskdecoder}) for generating the final mask is the same. Motivated by this, here we introduce contrastive learning that forces the network to narrow the distance of features of different expressions for the same target object while enlarging the distance of features for different objects. Furthermore, to provide more positive pairs and enhance the model's generalization ability to the input language, we randomly mask some specific words in the language expression and add these masked expressions to the positive samples of the original expression.

To sample the training pairs for contrastive learning, we summarize the inter-sample relationships into three categories: 1) Different Image (DI), 2) Same Image Different Object (SIDO), 3) Same Image Same Object (SISO), as shown in \figurename~\ref{fig:isl}. Unlike existing methods that construct the training batches in a fully random manner, we intend to let one batch have all kinds of inter-sample relationships. Firstly we randomly choose an initial sample $S_{init}$, as shown in \figurename~\ref{fig:isl_batch}. We denote its SISO images as $S_{SO}$, whose image $I$ and object $O_i$ are the same as $S_{init}$ but expressions $E_i^{j}$ are different from $S_{init}$, and denote its SIDO images as $S_{DO}$, which has the same image $I$ as the initial sample but different target object $O_i$. When constructing a mini-batch, we first put the initial sample into it. Next we intentionally put at most $N_{SO}$ samples from $S_{SO}$, and at most $N_{DO}$ samples from $S_{DO}$. The rest of the batch is filled with the randomly chosen DI samples. Under this mechanism, every training batch will contain all kinds of inter-sample relationships, as shown in \figurename~\ref{fig:isl_batch}.

As we mentioned earlier, the features of SISO samples for generating the final mask should be the same. In contrast, for SIDO items, though they share the same input image so the output feature of the transformer may tend to be similar, the features for generating the mask prediction should be different because their target outputs are different. From this point, we introduce contrastive learning as feature-level supervision. In our approach, the Mask Decoder module plays the role in generating the output mask, hence we add the contrastive learning on the feature $F_m$, see Eq.~(\ref{eq:maskdecoder}), of the Mask Decoder module. Inspired by the InfoNCE loss~\cite{van2018representation}, our loss is defined as follows:
\begin{equation}
\!\!\mathcal{L}_{CL} = \!-  \frac{1}{N_{SO}}\sum_{S_+ \in S_{SO}}\!\!\log \frac%
  {\exp\left(\frac{1}{\tau}\ \left< f_{S_+},f_{S_{init}}\right>\right)}%
  {\sum\limits_{S \in S_{DO}, S_+} \exp (\frac{1}{\tau}\left< f_{S},f_{S_{init}}\right>)}, %
  \label{Eq:8}
\end{equation}
where $S_+$ denotes a SISO sample of the initial sample, $\tau$ is a temperature constant, $f_S$ is the feature $F_m$ in the Mask Decoder module, and $\left<,\right>$ denotes the cosine similarity function. This loss function forces that the mask feature of the initial sample to be closer to its SISO samples that are supposed to have the identical output feature and mask, and force it to be away from its SIDO samples, which are supposed to have a non-overlap mask with it.

What's more, a sentence usually has more than one informative clue. However, the network tends to capture the most discriminative, or easiest clues to reach the training objective, resulting in underrating, even ignoring, other information. For example, given the image in \figurename~\ref{fig:isl} and an expression \textit{``man in green on left''}, we have experimentally observed that the network is over-influenced by the word \textit{``left''} since it is more common in the dataset. We argue that overly relying on discriminative/common words harms the model's generalization ability. To address this issue, we propose to randomly mask some prominent words in the language expression and add these masked samples, as SISO samples, to our contrastive learning. Specifically, we measure the importance of each word when evaluating the language attention in the QGM, as mentioned in Eq.~(\ref{eq:5}). For language expression that are longer than $N_m$ words, we sum up the word attention vectors of all the ${N_q}$ queries to generate a global attention weight for every words: $a_i = \sum_{n=1}^{N_q}{a_{ni}}$, where $a_{ni}\in A_\mathrm{qd}$ represents the attention weight of the $i$-th word in attention vector for the $n$-th query, $a_i$ denotes the global attention weight for $i$-th word. The global attention weights $\{a_1, a_2,...,a_{N_t}\}$ reveal the importance of each word. To enhance the diversity of the training samples, words are chosen to be masked with a probability $p_{m}$, where the probability $p_{m}$ is determined by the global attention weights by $p_{mi}=e^{a_i}/\sum^{N_t}_{j=1}e^{a_j}$. This probability-guided random setting lets more important words have higher masking chances while keeping the diversity of the training sentences. If a word is masked, its corresponding feature in $F_{t}$ is changed. Thus, we apply a softmax function in Eq.~(\ref{eq:5}) again to re-normalize it. As a consequence, a new set of query vectors and query responses are generated. The feature for Mask Decoder by this erased sentence is trained to be close to the original one by adding it as a positive sample $S_+$ in Eq.~(\ref{Eq:8}). In such a way, the network is encouraged to extract the information from words that are harder or not so discriminative rather than always relying on some high-frequency keywords, which could enhance the network's versatility in practical usage.

\subsection{Network Architecture}

\textbf{Feature Extractor.}~Since the transformer architecture only accepts sequential inputs, the original image, and language input must be transformed into feature space before sending to the transformer. For vision features, following~\cite{carion2020end}, we use a CNN backbone for image encoding. We take the features of the last three layers in the backbone as the input for our encoder. By resizing the three sets of feature maps to the same size and summing them together, we get the raw vision feature $F_{vr}\in \mathbb{R}^{H\times W \times C}$, where $H, W$ is the spatial size of features, and $C$ is the feature channel number. For language features, we first use a lookup table to convert each word into word embeddings~\cite{zhanghui2021}, and then utilize an RNN module to achieve contextual understanding of the input sentence and convert the word embedding to the same number of channels as the vision feature, resulting in a set of language features $F_t\in \mathbb{R}^{N_t\times C}$. $F_{vr}$ and $F_t$ are then sent to the Spatial Dynamic Fusion module and the Query Generation module as vision and language features.

\begin{table*}[t]
\renewcommand\arraystretch{1.0}
   \centering
   \small
   \caption{Comparison with Convolutional Networks, containing seven 3$\times$3 Conv layers, in terms of parameter size and performance.}
   \vspace{-0.1in}
   \setlength{\tabcolsep}{4.2mm}{\begin{tabular}{l|c|cccccc}
      \specialrule{.1em}{.05em}{.05em} 
      Type          & \#params     & IoU   &Pr@0.5 & Pr@0.6 & Pr@0.7 & Pr@0.8 & Pr@0.9 \\
      \hline \hline
      7 Conv Layers &$\sim 16.6$M & 60.42 & 66.44 & 61.86  & 53.22  & 44.72  & 17.27 \\
      Transformer   &$\sim 17.5$M & 65.24 & 73.39 & 68.01  & 60.83  & 47.99  & 20.07 \\
      \specialrule{.1em}{.05em}{.05em} 
   \end{tabular}}
   \label{tab:param}
\end{table*}

\textbf{Transformer Module.}~We use a complete but shallow transformer to apply the attention operations on input features. The network has a transformer encoder and a decoder, each of which has two layers. We use the standard Transformer architecture as defined by Vaswani \etal \cite{vaswani2017attention}, in which each encoder layer consists of one multi-head attention module and one feed-forward network (FFN), and each decoder layer consists of two multi-head attention modules and one FFN. We flatten the spatial domain of the fused multi-modal feature $F_{fused}$ into a sequence, forming the multi-modal feature $F_{fused}'\in \mathbb{R}^{N_v \times C}, N_v=HW$. The transformer encoder takes $F_{fused}'$ as input, deriving the memory features about vision information $F_{mem}\in \mathbb{R}^{N_v\times C}$. Before sending it to the encoder, we add a fixed sine spatial positional embedding on $F_{fused}'$. $F_{mem}$ is then sent to the transformer decoder as keys and values, together with $N_q$ query vectors produced by the Query Generation Module. The decoder queries the vision memory feature with language query vectors and outputs $N_q$ responses for mask decoding.

%-------------------------------------------------------------------------
\section{Experiments}
We conducted extensive experiments to demonstrate the effectiveness of our proposed Vision-Language Transformer (VLT) for referring segmentation. In this section, we introduce implementation details of our approach, benchmarks we used in the experiments, and report both the quantitative and qualitative results of our proposed approach compared with other state-of-the-art methods.

%-------------------------------------------------------------------------
\subsection{Implementation Details}
\textbf{Experiment Settings.}~Following previous works~\cite{luo2020multi,yu2018mattnet}, we use the same experiment settings. Our framework utilizes Darknet-53~\cite{yolo} pretrained on partial MSCOCO as the visual CNN backbone. Images form the validation and test set of the RefCOCO series are excluded in the pretraining. We use bi-GRU \cite{chung2014empirical} as the RNN implementation and the Glove Common Crawl 840B \cite{pennington2014glove} for word embedding. The training image size is set to $416\times 416$ pixels. Each Transformer block has eight heads, and the hidden layer size in all heads is set to 256. For RefCOCO and RefCOCO+, we set the maximum word number to 15, and for G-Ref, we set it to 20 as there are more long sentences. The Adam optimizer is used to train the network for 50 epochs, and the learning rate is set to $\lambda$~\!=~\!$0.001$. The batch size is 32 on one 32G V100 GPU.

\textbf{Metrics.} We use two metrics in our experiments: mask Intersection-over-Union (IoU) and Precision with thresholds (Pr@$X$). The mask IoU demonstrates the mask quality, which emphasizes the model's overall performance and reveals both targeting and segmenting abilities. The Pr@$X$ metric computes the ratio of successfully predicted samples using different IoU thresholds. Low threshold precision like Pr@0.5 reflects the identification performance of the method, and high threshold precision like Pr@0.9 reveals the ability of generating high-quality masks.

\begin{table*}[t]
\renewcommand\arraystretch{1.0}
   \centering
   \small
   \caption{Comparison of the proposed Query Generation Module (QGM) with other kinds of query generation ways. ``$F_t$'': directly use the language features $F_t$ as query vectors. ``Learnt'': learnable parameter-queries that are fixed in testing, similar with~\cite{carion2020end}.}
   \vspace{-0.1in}
   \setlength{\tabcolsep}{4.6mm}{\begin{tabular}{r|c|cccccc}
      \specialrule{.1em}{.05em}{.05em} 
      No. & Query Type  & IoU   & Pr@0.5 & Pr@0.6 & Pr@0.7 & Pr@0.8 & Pr@0.9 \\
      \hline \hline
      1   & $F_t$   & 60.26 & 69.88  & 64.61  & 56.70  & 43.62  & 18.06   \\
      2   & Learnt  & 58.60 & 67.84  & 59.98  & 53.23  & 44.60  & 16.33   \\
      \hline
      3   & QGM (ours) & 65.24 & 73.39 & 68.01  & 60.83  & 47.99  & 20.07 \\
      \specialrule{.1em}{.05em}{.05em} 
   \end{tabular}}
   \label{tab:exp_qgm}
\end{table*}

\begin{table}[t!]
\renewcommand\arraystretch{1.0}
   \centering
   \small
   \caption{Ablation study of Query Numbers $N_q$. $\ddagger$: without Query Balance Module (QBM).}
   \vspace{-0.1in}
    \setlength{\tabcolsep}{1.96mm}{
      \begin{tabular}{r|cccccc}
      \specialrule{.1em}{.05em}{.05em} 
      $N_q$ & IoU   & Pr@0.5 & Pr@0.6 & Pr@0.7 & Pr@0.8 & Pr@0.9 \\
      \hline \hline
      1     & 57.34 & 67.04 & 60.11 & 52.03 & 40.82 & 10.28 \\
      2     & 60.78 & 70.18 & 63.50  & 55.41 & 44.20  & 16.03 \\
      4     & 61.58 & 70.92 & 64.33 & 56.02 & 44.23 & 15.22 \\
      8     & 64.35 & 72.61 & 66.98 & 58.83 & 46.98 & 19.63 \\
      16    & \textbf{65.24} & \textbf{73.39} & 68.01 & 60.83 & 47.99 & 20.07 \\
      32    & 65.12 & 73.21 & 67.59 & 60.13 & 48.03 & 18.64 \\
      \hline
      16$^\ddagger$ & 63.80 & 71.96 & 67.46 & 59.73 & 47.22 & 19.71 \\
      \specialrule{.1em}{.05em}{.05em} 
      \end{tabular}
    }
   \label{tab:exp_nq}%
\end{table}%

%-------------------------------------------------------------------------
\subsection{Datasets}\label{sec:dataset}

The proposed VLT is evaluated on three public referring segmentation datasets: RefCOCO, RefCOCO+~\cite{yu2016modeling} and G-Ref~\cite{mao2016generation, nagaraja2016modeling}.

\textbf{RefCOCO \& RefCOCO+}~\cite{yu2016modeling} are two of the largest image datasets for referring segmentation. They are also called UNC \& UNC+ datasets in some literature. 142,209 referring language expressions describing 50,000 objects in 19,992 images are collected in the RefCOCO dataset, and 141,564 referring language expressions for 49,856 objects in 19,992 images are collected in the RefCOCO+ dataset.
The difference between two datasets is that the RefCOCO+ restricts the expression ways for the language sentences. For example, descriptions about absolute locations, \eg, \textit{``leftmost''}, are forbidden in the RefCOCO+ dataset.

\textbf{G-Ref}~\cite{mao2016generation, nagaraja2016modeling}. Also called RefCOCOg, it is another famouse and well recognized referring segmentation dataset. 104,560 referring language expressions for 54,822 objects in 26,711 images are used in G-Ref. Unlike RefCOCO \& RefCOCO+, the language usage in the G-Ref is more casual but complex, and the sentence lengthes of G-Ref are also longer in average. Notably, G-Ref has two versions: one is called UMD split~\cite{nagaraja2016modeling}, the other is called Google split~\cite{mao2016generation}. The UMD split has both validation and testing set publicly available, but the Google split only makes its validation set public. We report the results of the proposed VLT on both UMD and Google version.

%-------------------------------------------------------------------------
\subsection{Ablation Study}

In this section, we conduct ablation studies on the {test B of RefCOCO} to demonstrate the effectiveness of the proposed modules in our Vision-Language Transformer framework.

\textbf{Transformer \vs ConvNet.} To demonstrate the scale of our proposed network and verify the effectiveness of the transformer module, we compare our method with a regular ConvNet in terms of the performance and parameter size in \tablename~\ref{tab:param}.
In the experiment, we replace the whole transformer-based modules, including the transformer encoder-decoder, the Query Generation Module, and the Query Balance Module with seven stacked $3\times 3$ Conv layers that have similar parameters size to our transformer-based modules. It shows that the parameter size of our transformer-based module achieves a much superior performance while is only nearly equal to 7 convolutional layers. The transformer module outperforms the 7 Conv module with {$\sim$5\% margin in terms of IoU, and $\sim$7\% margin in terms of Precision@0.5}. This proves the effectiveness of the proposed transformer module.

\textbf{Query Generation.} In \tablename~\ref{tab:exp_qgm}, we compare different kinds of query generation methods, including our proposed Query Generation Module (QGM), language features $F_t$ as queries, and learned parameters as queries. The Query Generation Module outperforms the other two methods with a large margin at about {5\% - 7\%} in terms of IoU and 4\% - 6\% in terms of Pr@0.5. Firstly, we directly utilize the language features $F_t$ as query vectors and send them into the transformer decoder. In detail, the given language expression is processed by an RNN network, then the output for every word, and the output for the whole sentence, are used as query vectors. It can be seen in \tablename~\ref{tab:exp_qgm} that the performance of $F_t$ as queries is $\sim$5\% worse than QGM, which is because the information between words is not sufficiently exchanged and the understanding of language is derived from language itself, as we discussed in Sec~\ref{sec:QGM}. The proposed Query Generation Module has a much superior performance to the ``$F_t$'' as queries. This demonstrates that the proposed QGM effectively understands the language expressions and produces valid attended language features under the guidance of visual information. We set 16 query vectors that are initialized with uniform distribution at the beginning of the training in our experiment, and train these query-parameters together with the network. As the ``learnt'' in \tablename~\ref{tab:exp_qgm}, the performance of these learned fixed query vectors is not satisfying, only 58.50\%, which shows that such learned query-parameters cannot represent the target object as effectively as online generated input-specific queries by the proposed QGM. 

\begin{figure}[t]
   \begin{center}
      \includegraphics[width=\linewidth]{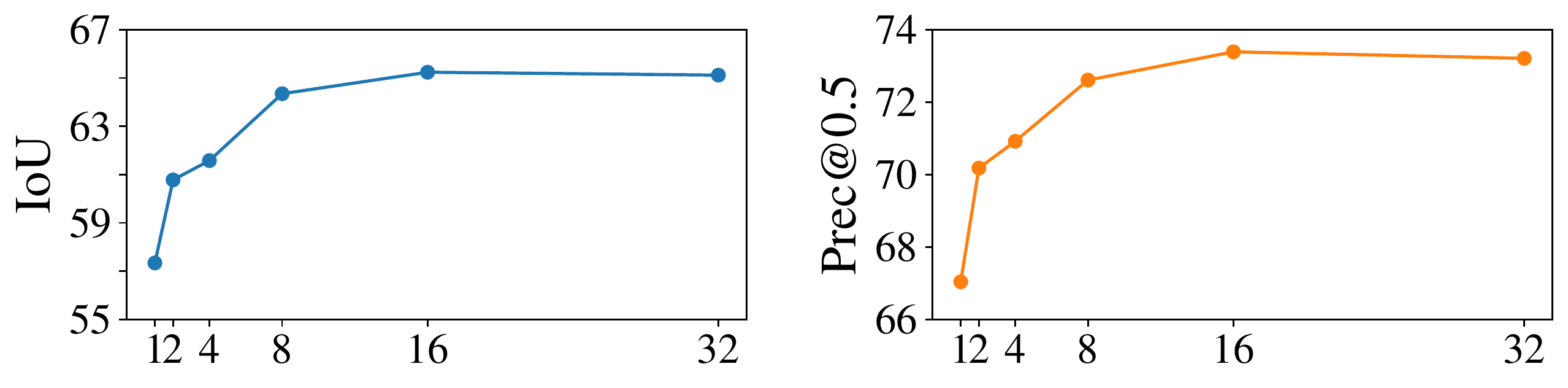}
   \end{center}
   \vspace{-0.1in}
   \caption{Performance gain by increasing the query number $N_q$.}
   \vspace{-0.1in}
   \label{fig:nq}
\end{figure}
\textbf{Query Number $N_q$.} To demonstrate the influence of the query number $N_q$ on the results, we evaluate the network's results with different numbers of query vectors. As we can see in \tablename~\ref{tab:exp_nq} and \figurename~\ref{fig:nq}, though only one segmentation mask is required in the final prediction, multiple queries are desired for providing diverse clues and can achieve better results than a single query. As shown in \tablename~\ref{tab:exp_nq} and \figurename~\ref{fig:nq}, by increasing the query number $N_q$, the performance gradually gets higher, and a significant performance gain of about {8\% is achieved from 1 query to 16 queries}. 
The performance gain slows down after the query number $N_q$ is larger than 8, therefore we select $N_q$ = 16 as the default setting. The performance gain achieved by larger $N_q$ verifies that multiple input-specific queries produced by the proposed QGM dynamically represent the diverse comprehensions of language expression. When the Query Balance Module (QBM) is discarded, marked with $\ddagger$ in \tablename~\ref{tab:exp_nq}, {a performance drop of 1.44\% IoU is observed, which proves the advantage of the proposed QBM.}

\begin{table}[t!]
\renewcommand\arraystretch{1.0}
   \centering
   \small
   \caption{Ablation study of Multi-Modal Fusion.}
   \vspace{-0.1in}
   \setlength{\tabcolsep}{0.8mm}{\begin{tabular}{c|cccccc}
      \specialrule{.1em}{.05em}{.05em} 
      Type & IoU   & Pr@0.5 & Pr@0.6 & Pr@0.7 & Pr@0.8 & Pr@0.9 \\
      \hline \hline
      Tile & 64.40&  72.16 & 66.82  & 58.33  & 47.20  & 20.03  \\
      Tile\scriptsize{~+~Conv$\times$4} & 64.45&  72.19 & {66.63}  & {58.23}  & {47.40}  & {20.01}  \\
      SDF  & 65.24 & 73.39 & 68.01  & 60.83  & 47.99  & 20.07 \\
      \specialrule{.1em}{.05em}{.05em} 
   \end{tabular}}
   \label{tab:SDF}
\end{table}

\textbf{Tile \vs Spatial-Dynamic Fusion.} In \tablename~\ref{tab:SDF}, we compare the ``tile-and-concatenate'' fusion and our proposed Spatial-Dynamic Fusion (SDF). As we discussed in Section~\ref{sec:sdf}, the ``tile'' operation does not consider the difference of each pixel but uses an identical sentence feature for all pixels across the image. In contrast, the proposed spatial-dynamic fusion customizes a unique language feature for every pixel according to the interaction between language information and corresponding pixel information. As shown in \tablename~\ref{tab:SDF}, compared with ``Tile'', the SDF module brings a performance gain of {0.84\% IoU and 1.23\% Pr@0.5.} The proposed SDF emphasizes the differences among pixels/objects and allows each position to select the more informative words, enhancing the multi-modal fusion and producing better multi-modal features. ``Tile~+~Conv$\times4$'' in \tablename~\ref{tab:SDF}, which has the same number of parameters as the proposed SDF, does not bring better performance than ``Tile'' because our network already has a sequential convolution layers after the feature fusion.

\begin{table}[t]
   \renewcommand\arraystretch{1.0}
   \centering
   \small
   \caption{Ablation study of Inter-Sample Learning.}
   \begin{subtable}{\linewidth}
      \caption{Experiments on original dataset and masked dataset}
      \vspace{-0.076in}
      \setlength{\tabcolsep}{1.5mm}{\begin{tabular}{l|ccc|ccc}
            \specialrule{.1em}{.05em}{.05em}
            \multirow{2}{*}{Type} & \multicolumn{3}{c|}{IoU} & \multicolumn{3}{c}{Pr@0.5} \\
             & {Original} & {Masked} & {Gap} & {Original} & {Masked} & {Gap} \\
            \hline\hline
            w/o~~CL & 63.43& 59.53& -3.90 & 71.84 & 67.02 & -4.82 \\
            w/~~~~CL  & 64.70& 61.02& -3.68 & 72.88 & 68.45 & -4.43 \\
            w/~MCL & 65.24& 64.20& -1.04 & 73.39 & 72.19 & -1.20 \\
            \specialrule{.1em}{.05em}{.05em}
         \end{tabular}}
      \label{tab:maskedset}
   \end{subtable}
   \vspace{0.06in}

   \begin{subtable}{\linewidth}
      \caption{Cross dataset validation}
      \vspace{-0.076in}
      \setlength{\tabcolsep}{1.4mm}{\begin{tabular}{l|cccccc}
            \specialrule{.1em}{.05em}{.05em}
            Type               & IoU   & Pr@0.5 & Pr@0.6 & Pr@0.7 & Pr@0.8 & Pr@0.9 \\
            \hline\hline
            w/o~~\!~CL$^\dagger$  & 49.16 & 56.06  & 50.13  & 41.87  & 34.11  & 12.26  \\
            w/~~~~CL$^\dagger$ & 49.92 & 57.01  & 51.25  & 42.13  & 35.54  & 12.81  \\
            w/~MCL$^\dagger$   & 52.35 & 60.41  & 55.12  & 49.80  & 39.35  & 14.76  \\
            Native             & 56.30 & 66.03  & 61.53  & 56.20  & 41.22  & 13.09  \\
            \specialrule{.1em}{.05em}{.05em}
         \end{tabular}}
         \label{tab:crossset}

   \end{subtable}
   \label{tab:inter_sample_learning}%

\end{table}%
 
\begin{table}[t]
   \renewcommand\arraystretch{1.0}
   \centering
   \small
  \caption{Ablation study of word selection mechanism in MCL.}
  \begin{subtable}{\linewidth}
  \centering
      \vspace{-0.06in}
      \setlength{\tabcolsep}{1.2mm}{\begin{tabular}{c|cccccc}
            \specialrule{.1em}{.05em}{.05em}
            Select. Type               & IoU   & Pr@0.5 & Pr@0.6 & Pr@0.7 & Pr@0.8 & Pr@0.9 \\
            \hline\hline
            None            & 49.92 & 57.01  & 51.25  & 42.13  & 35.54  & 12.81  \\
            Random             & 50.08 & 57.33  & 51.30  & 43.02  & 35.72  & 12.60  \\
            Threshold $\theta$ & 51.57 & 59.08  & 52.04  & 46.13  & 37.57  & 13.99  \\
            $p_{m}$            & \textbf{52.35} & \textbf{60.41}  & \textbf{55.12}  & \textbf{49.80}  & \textbf{39.35}  & \textbf{14.76}  \\
            \specialrule{.1em}{.05em}{.05em}
         \end{tabular}}
  \end{subtable}
   \label{tab:wordselection}
\end{table}%

\begin{table*}[t]
\renewcommand\arraystretch{1.0} 
   \centering
   \small
   \caption{Results on Referring Image Segmentation in terms of IoU and Prec@0.5. U: UMD split. G: Google split. Methods pretrained on large-scale vision-language training datasets are marked with $\dagger$.}
  \vspace{-0.1in}
   \setlength{\tabcolsep}{2.4mm}{\begin{tabu}{l|c|c|ccc|ccc|ccc}
      \specialrule{.1em}{.05em}{.05em} 
      \multirow{2}{*}{Methods} &\multirow{2}{*}{\shortstack{Visual\\Backbone}} & \multirow{2}{*}{\shortstack{Textual\\Encoder}}&\multicolumn{3}{c|}{RefCOCO} & \multicolumn{3}{c|}{RefCOCO+} & \multicolumn{3}{c}{G-Ref} \\
    %   \cline{3-11}
                              &&& val   & test A & test B & val   & test A & test B & val$_\text{(U)}$   & test$_\text{(U)}$  & val$_\text{(G)}$\\
      \hline
      \hline
      DMN~\cite{margffoy2018dynamic} &DPN92& SRU &49.78 & 54.83 & 45.13 & 38.88 & 44.22 & 32.29 & -     & -     & 36.76 \\
      RRN~\cite{li2018referring}     &DeepLab-R101& LSTM &55.33 & 57.26 & 53.93 & 39.75 & 42.15 & 36.11 & -     & -     & 36.45 \\
      MAttNet~\cite{yu2018mattnet}   &MaskRCNN-R101& bi-LSTM &56.51 & 62.37 & 51.70 & 46.67 & 52.39 & 40.08 & 47.64 & 48.61 & -     \\
      CMSA~\cite{ye2019cross}        &DeepLab-R101& None &58.32 & 60.61 & 55.09 & 43.76 & 47.60 & 37.89 & -     & -     & 39.98 \\
      CAC~\cite{Chen_lang2seg_2019}  &ResNet101& bi-LSTM &58.90 & 61.77 & 53.81 & -     & -     & -     & 46.37 & 46.95 & 44.32 \\
      STEP~\cite{chen2019see}        &DeepLab-R101& bi-LSTM &60.04 & 63.46 & 57.97 & 48.19 & 52.33 & 40.41 & -     & -     & 46.40 \\
      BRINet~\cite{hu2020bi}         &DeepLab-R101& LSTM &60.98 & 62.99 & 59.21 & 48.17 & 52.32 & 42.11 & -     & -     & 48.04 \\
      CMPC~\cite{huang2020referring} &DeepLab-R101& LSTM &61.36 & 64.53 & 59.64 & 49.56 & 53.44 & 43.23 & -     & -     & 39.98 \\
      LSCM~\cite{hui2020linguistic}  &DeepLab-R101& LSTM &61.47 & 64.99 & 59.55 & 49.34 & 53.12 & 43.50 & -     & -     & 48.05 \\
      CMPC+~\cite{liu2021crossTPAMI} &DeepLab-R101& LSTM &62.47 & 65.08 & 60.82 & 50.25 & 54.04 & 43.47 & -     & -     & 49.89 \\
      MCN~\cite{luo2020multi}        &Darknet53& bi-GRU &62.44 & 64.20 & 59.71 & 50.62 & 54.99 & 44.69 & 49.22 & 49.40 & -     \\
      EFN~\cite{feng2021encoder}     &ResNet101& bi-GRU &62.76 & 65.69 & 59.67 & 51.50 & 55.24 & 43.01 & -     & -     & 51.93 \\
      BUSNet~\cite{yang2021bottom}   &DeepLab-R101& Self-Att &63.27 & 66.41 & 61.39 & 51.76 & 56.87 & 44.13 & -     & -     & 50.56 \\
      CGAN~\cite{luo2020cascade}     &DeepLab-R101& bi-GRU &64.86 & 68.04 & 62.07 & 51.03 & 55.51 & 44.06 & 51.01 & 51.69 & 46.54 \\
      ISFP~\cite{liu2022instance}&Darknet53&Bi-GRU&{65.19} & {68.45} & {62.73} & {52.70} & {56.77} & {46.39} & {52.67} & {53.00} & 50.08 \\
      LTS~\cite{jing2021locate}      &Darknet53& bi-GRU &65.43 & 67.76 & 63.08 & 54.21 & 58.32 & 48.02 & 54.40 & 54.25 & -     \\
      \hline
      \textbf{VLT~(ours)} &Darknet53& bi-GRU &\textbf{67.52} & \textbf{70.47} & \textbf{65.24} & \textbf{56.30} & \textbf{60.98} & \textbf{50.08} & \textbf{54.96} & \textbf{57.73} & \textbf{52.02} \\
      \hline\hline
      {ReSTR}~\cite{kim2022restr} &ViT-B& Transformer &67.22 & 69.30 & 64.45 & 55.78 & 60.44 & 48.27 & -     & -     & 54.48 \\
      {MaIL}~\cite{MaIL}$^\dagger$ & {ViLT}& BERT &70.13& 71.71 & 66.92 & 62.23 & 65.92 & 56.06 & 62.45 & 62.87     &61.81\\
      {CRIS}~\cite{wang2022cris}$^\dagger$& CLIP-R101& CLIP &70.47 & 73.18 & 66.10 & 62.27 & 68.08 & 53.68 & 59.87     & 60.36     & - \\
      {LAVT}~\cite{yang2021lavt}&{Swin-B}& BERT &72.73 & {75.82} & 68.79 & 62.14 & 68.38 & 55.10 & 61.24     & 62.09     & 60.50 \\
      \hline
      {\textbf{VLT~(ours)}} &{Swin-B}& BERT&\textbf{72.96} & \textbf{75.96} & \textbf{69.60} & \textbf{63.53} & \textbf{68.43} & \textbf{56.92} & \textbf{63.49} & \textbf{66.22} & \textbf{62.80} \\
      \hline
      \multicolumn{3}{l|}{VLT$_\text{Darknet53}$~(ours) Prec@0.5} & 77.03 & 81.01 & 73.39 & 66.03 & 71.87 & 56.91 & 62.05 & 60.96 & 57.88 \\
      \multicolumn{3}{l|}{VLT$_\text{Swin-B}$~~~~(ours) Prec@0.5} & 85.35 & 87.76 & 80.48 & 74.95 & 80.98 & 67.44 & 77.23 & 78.03 & 73.84 \\
      \specialrule{.1em}{.05em}{.05em} 
   \end{tabu}}%
  \vspace{-0.1in}
   \label{tab:results}%
\end{table*}%

\begin{figure}[t]
   \begin{center}
      \includegraphics[width=1\linewidth]{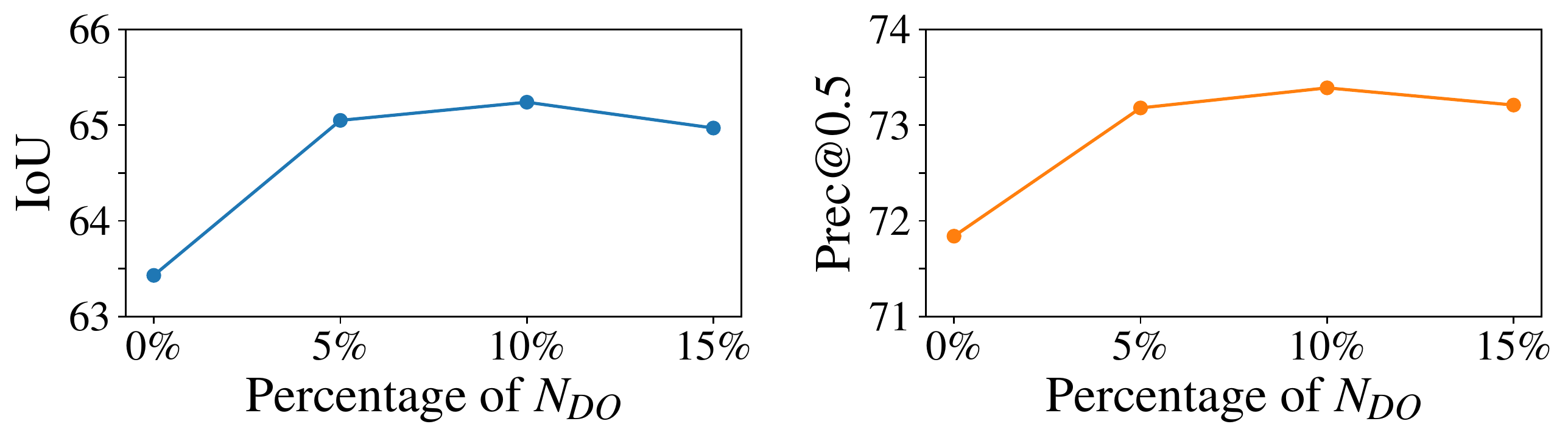}
   \end{center}
   \vspace{-0.1in}
   \caption{Ablation study of the percentage of $N_{DO}$ in MCL.}
   \label{fig:nso}
\end{figure}
\textbf{Inter-Sample Learning.} Here we demonstrate the effectiveness of our proposed inter-sample learning approach, Masked Contrastive Learning (MCL). The results are shown in \tablename~\ref{tab:inter_sample_learning}. Firstly, we add Contrastive Learning (CL) in the training of our network. The CL does not contain masked sentences as SISO samples. From \tablename~\ref{tab:maskedset}, on the original testing set, the CL brings a performance gain of {1.27\% in terms of IoU and 1.04\% in terms of Pr@0.5}, which demonstrates that the inter-sample learning does enhance the model's performance. Further, we introduce the samples with masked sentences as SISO samples, \ie, positive pairs in contrastive learning. Compared with w/o CL, MCL brings a large performance gain of 1.81\% IoU on the original dataset. Compared with CL, MCL further brings a performance gain of 0.54\% in terms of IoU and 0.51\% in terms of Pr@0.5, which shows the benefits brought by introducing samples with masked expressions in training. 
To better demonstrate the model's ability in dealing with unconstrained and diverse language expressions in the wild, we do another two testings: 1) erase some informative words of the given language expressions in these testing samples, see ``Masked'' in \tablename~\ref{tab:maskedset}; 2) cross datasets validation between two datasets that have different common clues, \ie, training on RefCOCO while testing on the validation set of RefCOCO+, marked with w/o CL$^\dagger$, \etc in \tablename~\ref{tab:crossset}. Firstly, as shown in \tablename~\ref{tab:maskedset}, compared with the original dataset, w/o CL drops 3.9\% in terms of IoU and 4.82\% in terms of Pr@0.5 on the Masked testing samples. The result shows that the w/o CL model overly relies on common keywords and is heavily affected by the missing of these common clues. While for w/ MCL, the performance drop on the Masked validation is {1.04\% and is much less than the performance drop of w/o CL}, which verifies the model's robustness and generalization ability brought by introducing masked contrastive learning. 
Next, we do the cross-dataset validation on RefCOCO and RefCOCO+ in \tablename~\ref{tab:crossset}. In RefCOCO, a large number of samples use absolute location (\eg, \textit{``the left''}, \textit{``on the right''}, \etc) for describing the target object, but such kinds of expressions are not allowed in the RefCOCO+. Therefore, the cross datasets validation provides a good simulation of a practical scenario, in which the training information and testing are inconsistent, and only partial clues are available for testing. As shown in \tablename~\ref{tab:crossset}, w/ MCL$^\dagger$ outperforms w/o CL$^\dagger$ 3.19\% in terms of IoU and 4.35\% in terms of Pr@0.5, which verifies the model's robustness and generalization ability in dealing with diverse language expressions that are different from training samples. ``Native'' in \tablename~\ref{tab:crossset} denotes training \& testing on RefCOCO+. As w/ MCL$^\dagger$ \vs ``Native'', we can see that the model trained on RefCOCO with MCL achieves competitive results on the validation set of RefCOCO+ compared to the model trained on RefCOCO+, proving that the proposed masked contrastive learning enhances the model's generalizability under open-world practical scenarios.

Next we do an ablation study about the word selection mechanism in our masked constrastive learning. Apart from the baseline model that disables MCL, we test three mask-word selection methods: 1) randomly choose a word to mask, 2) randomly choose a word with the weight $a_i$ greater than a threshold $\theta$ to mask, 3) the proposed method that words are masked based on the probability $p_{m}$. \tablename~\ref{tab:wordselection} shows that our method outperforms other mask word selection mechanisms.

For the setting of $N_{DO}$ in MCL, the ablation study in \figurename~{\ref{fig:nso}} shows that the performance of the network reaches the peak when $N_{DO}$ is set to $10\%$ of the batch size. For $N_{SO}$, as the average number of expressions for an object is around 3, we can include all available Same Object (SO) samples in most cases.

\begin{figure}[t]
   \begin{center}
      \includegraphics[width=0.96\linewidth]{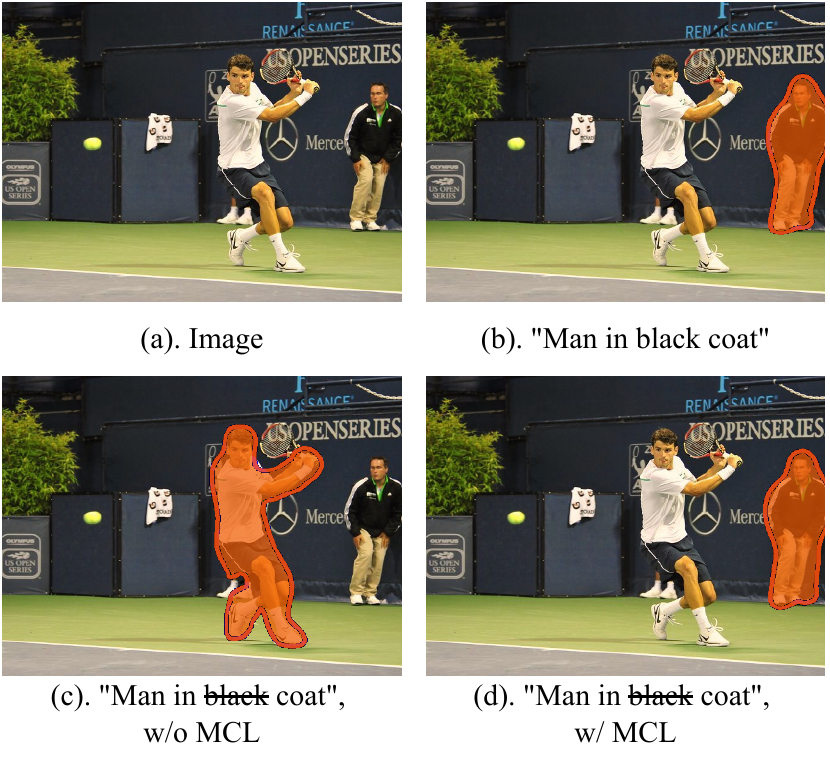}
   \end{center}
   \vspace{-0.1in}
   \caption{Example results of the Masked Contrastive Learning.}
   \label{fig:mcl_ablation}
\end{figure}

In \figurename~\ref{fig:mcl_ablation}, we provide qualitative examples to show the effectiveness of the Masked Contrastive Learning. The original input language expression contains information in two aspects: color (\textit{``black''}) and attribute (\textit{``coat''}). The model without MCL overly relies on the more obvious color information (\textit{``black''}), so it fails to predict when the word is erased. In contrast, the model with MCL successfully finds the target with partial information, showing that the MCL enhances the model's generalization ability to various language expressions.

We further test the training efficiency of our mask contrastive learning approach. We train the network with and without the MCL and report the GPU memory usage during training and the training speed of two runs with batch size set to 16. With MCL enabled, the GPU memory usage and average training speed are 18496MB and 0.479s/iter, respectively. Without MCL, they are 17842MB and 0.471s/iter, respectively. The increased training memory and time by MCL are less than $4\%$ and $2\%$, respectively.

%-------------------------------------------------------------------------
\subsection{Comparison with State-of-the-art Methods}

Here we compare the proposed Vision-Language Transformer (VLT) framework with previous state-of-the-art referring image segmentation methods on three commonly-used benchmarks, RefCOCO, RefCOCO+, and G-Ref. The results are reported in \tablename~\ref{tab:results}. It can be seen that the proposed VLT outperforms previous state-of-the-art methods on all three benchmarks. On RefCOCO, the IoU performance of the proposed VLT is better than other methods, \eg, LTS~\cite{jing2021locate}, with {$\sim$2\% gain} on three different testing splits. Then on RefCOCO+, the proposed VLT achieves new state-of-the-art result and is {around 2\% better than previous state-of-the-art method}. On the hard benchmark G-Ref that has longer language expressions, the proposed VLT consistently achieves new state-of-the-art referring segmentation performance with {an IoU improvement of about 0.5\%-3\%}, which demonstrates that the proposed VLT has good abilities to deal with hard cases and long expressions. We assume the reason is that, on the one hand, long and complex expressions usually contain more clues and more emphasis, and our proposed Query Generation Module and Query Balance Module can produce multiple comprehensions with different emphases and find the more suitable ones. On the other hand, harder cases also contain complex scenarios that need a holistic view and understanding of the given language expression and image, and the multi-head attention is more appropriate for such complex scenarios as a global operator. Meantime, compared with other methods that with stronger backbones, \eg, DeepLab-R101~\cite{chen2017deeplab}, MaskRCNN-R101~\cite{he2017mask}, ResNet101~\cite{he2016deep}, our backbone Darknet53 and our proposed modules are lightweight.

To compare with methods using stronger backbones, we further provide results with stronger visual and textual encoders in  \tablename~\ref{tab:results}. We use the popular vision transformer backbone Swin-B~\cite{liu2021Swin} as visual encoder and BERT~\cite{devlin2018bert} as textual encoder to replace the Darknet53~\cite{yolo} and bi-GRU~\cite{chung2014empirical}, respectively. Methods pretrained on large-scale vision-language datasets are marked with $\dagger$, \eg, MaIL~\cite{MaIL} adopts ViLT~\cite{kim2021vilt} pre-trained on four large-scale vision-language pretraining datasets and CRIS~\cite{wang2022cris} employs CLIP~\cite{radford2021learning} pretrained on 400M image-text pairs. As shown in \tablename~\ref{tab:results}, the proposed approach outperforms MaIL and CRIS by around $2\%\!\sim\!4\%$ IoU without using large-scale vision-language datasets in pretraining, which demonstrates the effectiveness of our proposed modules with stronger visual and textual encoders. Especially, the proposed approach VLT achieves higher performance gain on more difficult dataset G-Ref that has a longer average sentence length and more complex and diverse word usages, \eg, VLT is $\sim\!\!4\%$ IoU better than MaIL~\cite{MaIL} and LAVT~\cite{yang2021lavt} on test$_\text{(U)}$ of G-Ref. It demonstrates the proposed model's good ability in dealing with long and complex expressions with large diversities, which is mainly attributed to input-conditional query generation and selection that well cope with the diverse words/expressions, and masked contrastive learning that enhances the model's generalization ability.

%-------------------------------------------------------------------------
\subsection{Qualitative Results and Visualization}

\begin{figure}[t]
   \centering
   \begin{subfigure}[b]{0.45\linewidth}
       \centering
       \includegraphics[height=0.76in]{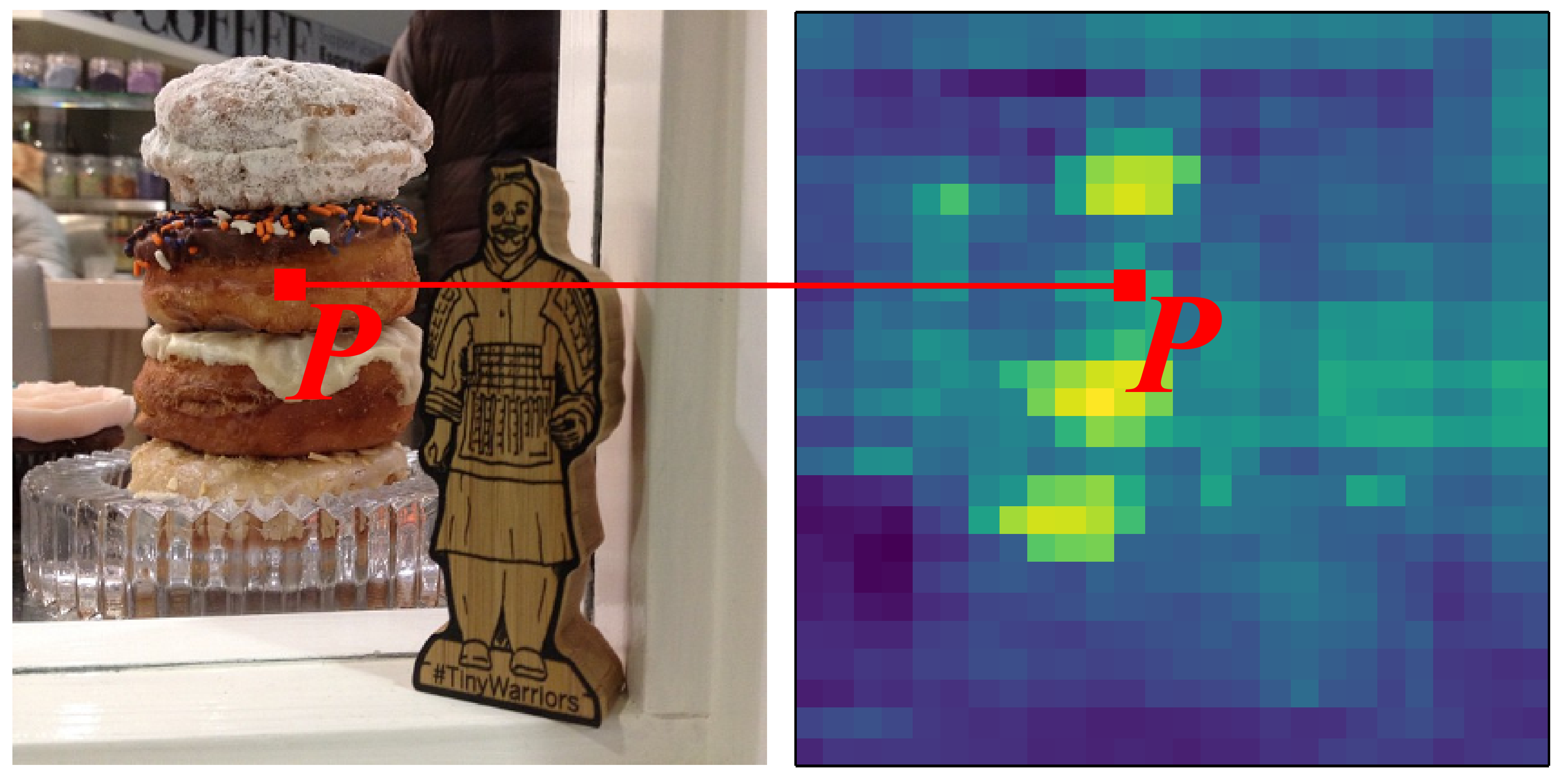}
       \caption{}
       \label{fig:visA}
   \end{subfigure}
   \hfill
   \begin{subfigure}[b]{0.5\linewidth}
       \centering
       \includegraphics[height=0.762in]{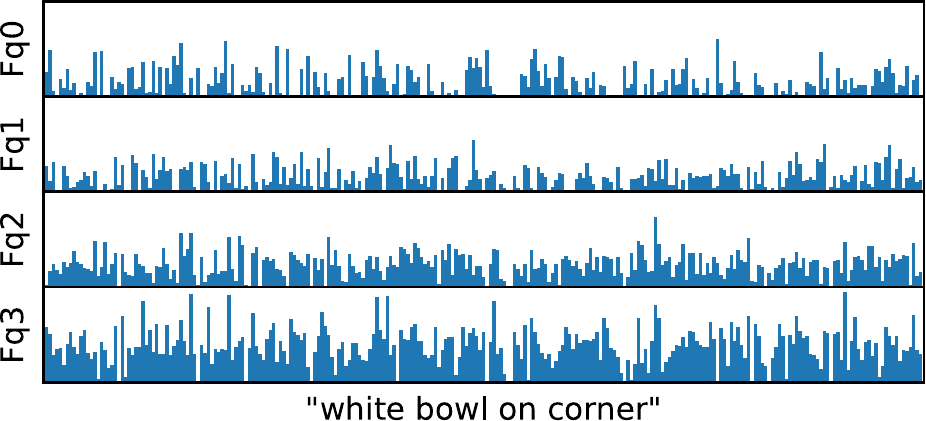}
       \caption{}
       \label{fig:visQ}
   \end{subfigure}
   \vspace{-0.1in}
   \caption{Visualizations of: (a) the attention map of point $P$ in the transformer encoder; (b) different query vectors $F_q$.}
   \vspace{-0.1in}
\end{figure}

In \figurename~\ref{fig:visA}, we extract and visualize an attention map for a position ``$P$'' from the 2nd layer of our transformer encoder. It shows that in a single layer of the transformer, the attention of one output pixel globally extends to other input pixels far away.
We also see that pixel on one instance attends to other instances, showing our network is able to capture long-range interactions between instances.
In \figurename~\ref{fig:visQ}, we visualize four query vectors $F_q$ (see \figurename~\ref{fig:query_gen} and Eq.~(\ref{Eq:Fq})). The four query vectors differ from each other and have different distributions of response peaks, which demonstrates the diversity of these input-specific query vectors.

\begin{figure*}[htbp]
   \begin{center}
      \includegraphics[width=1\linewidth]{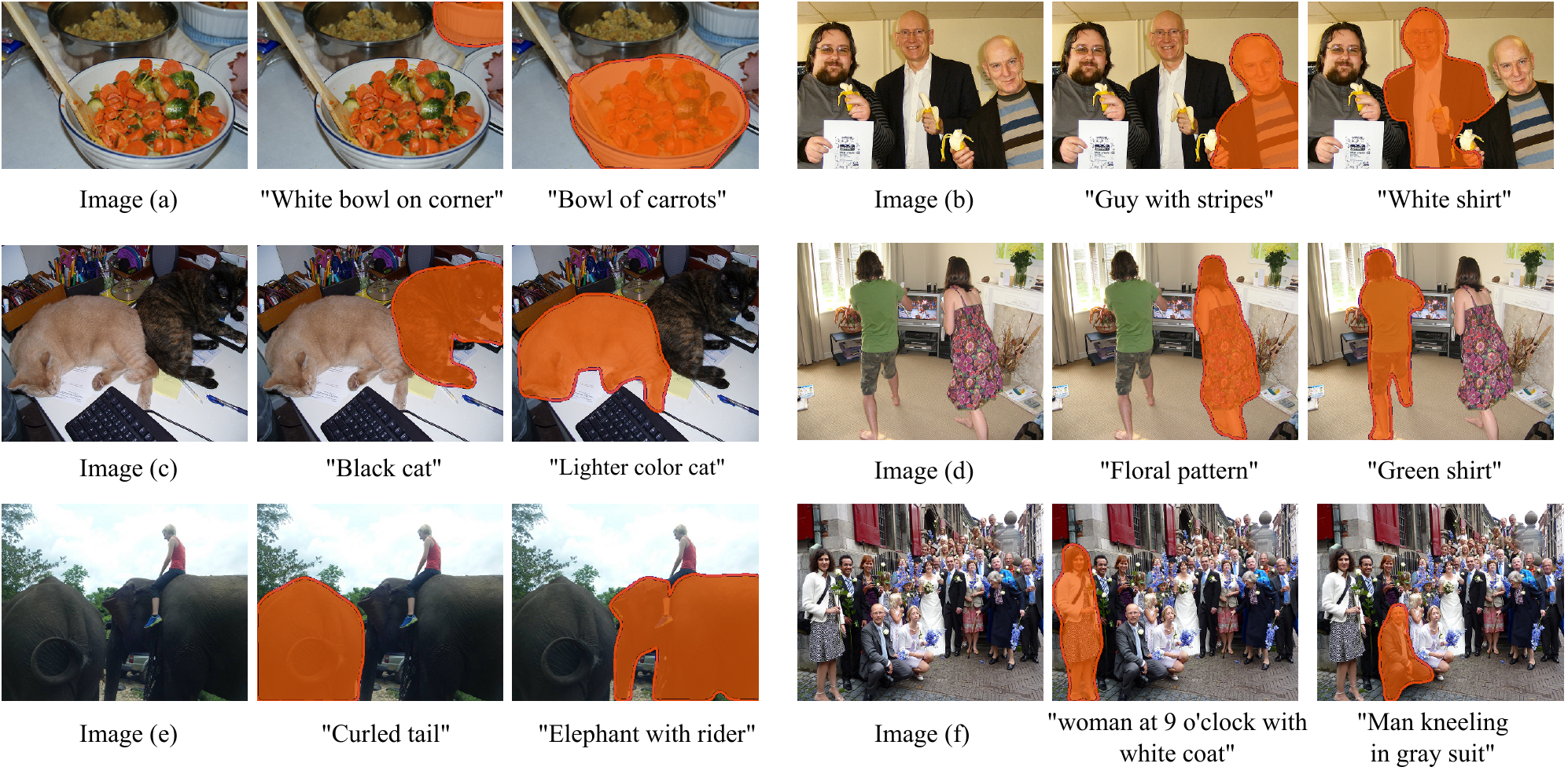}
   \end{center}
  \vspace{-0.16in}
   \caption{(Best viewed in color) Qualitative examples of the proposed VLT. For each example, the first image is the input image, and captions under the second and third images are the given language expressions.}
   \label{fig:demo}
\end{figure*}

Then, we visualize some qualitative examples of the proposed VLT in \figurename~\ref{fig:demo}. To demonstrate the identifying ability of our VLT, we show the mask predictions of two different input language expressions for every example. Image (a) and (c) are two typical examples that the language expression directly provides the location or color clues of the target object. In the second expression of Image (c), \textit{``lighter color cat''}, it can be seen that the proposed VLT is able to handle the expressions that indicate the target object by providing a comparison of it with other objects, \eg, \textit{``lighter''}. The examples of image (b) and (d) demonstrate the model's ability on understanding the attribute words, \eg, \textit{``stripes''}, and relatively rarer words, \eg, \textit{``floral''}. In the second expression of image (e), our VLT successfully identifies the target object referred by expression describing the relationships between objects, \ie, ``Elephant with rider''. Image (f) contains a group of people, where all instances distribute densely in a complicated layout. The proposed method manages to identify the target instance with difficult language expressions that contain multiple aspects of clues, such as direction (\textit{``9 o'clock''}), attributes (\textit{``white coat''} \& \textit{``gray suit''}), and posture (\textit{``kneeling''}).

\subsection{Results on Referring Video Object Segmentation}

Our proposed approach can also be applied to referring video object segmentation (RVOS) task with minor adaptation. We apply our model on each individual frame of the input video clip. We use the average vision features of all frames of a video clip as the vision features in the QGM ($F_{vq}$ in \figurename~{\ref{fig:query_gen}}). This enables the query input to be kept identical for all frames in a video clip, achieving a temporal consistency across frames. When performing the contrastive learning on video data, we sample different objects $S_{DO}$ in the $\pm 2$ adjacent frames of the initial object. As adjacent frames shares similar image structure with the original frame, we can enlarge the number of negative samples while keeping a similar behavior with our image model. According to the experiments, when only sampling $S_{DO}$ in the same video frame, the $\mathcal{J}\&\mathcal{F}$ performance is 63.5 while it increases to 63.8 when adding the $\pm 2$ adjacent frames.

\begin{table}[t]
   \renewcommand\arraystretch{1.06}
   \centering
   \small
   \caption{Results on Referring Video Object Segmentation.}
  \vspace{-0.1in}
   \setlength{\tabcolsep}{0.6mm}{\begin{tabular}{l|c|ccc|ccc}
         \specialrule{.1em}{.05em}{.05em}
         {\multirow{2}[0]{*}{Methods}} & \multirow{2}{*}{Backbone}&\multicolumn{3}{c|}{YouTube-RVOS} & \multicolumn{3}{c}{Ref-DAVIS17} \\
         {} && $\mathcal{J\&F}$ & $\mathcal{J}$ & $\mathcal{F}$ & $\mathcal{J\&F}$ & $\mathcal{J}$ & $\mathcal{F}$ \\
         \hline \hline
         CMSA~\cite{ye2019cross} & ResNet50 & 36.4 & 34.8 & 38.1 & 40.2 & 36.9 & 43.5 \\ 
         URVOS~\cite{seo2020urvos} & ResNet50 & 47.2 & 45.3 & 49.2 & 51.5 & 47.3 & 56.0 \\
         PMINet~\cite{ding2021pminet} & ResNeSt101 & 53.0 & 51.5 & 54.5 & - & - & - \\
         CITD~\cite{liang2021topdown} & Ensemble & 61.4 & 60.0 & 62.7 & - & - & - \\
         ReferFormer~\cite{wu2022referformer} & V-Swin-B&62.9 & 61.3 & 64.6 & 61.1 & 58.1 & 64.1 \\
         \hline
         \textbf{VLT~(ours)}       & V-Swin-B&\textbf{63.8} & \textbf{61.9} & \textbf{65.6} & \textbf{61.6} & \textbf{58.9} & \textbf{64.3} \\
         \specialrule{.1em}{.05em}{.05em}
      \end{tabular}}%
   \label{tab:rvos}%
   \vspace{-0.1in}
\end{table}%

In \tablename~{\ref{tab:rvos}}, we report the quantitative results of the proposed VLT on the validation set of the YouTube-RVOS~\cite{seo2020urvos} dataset and Ref-DAVIS17~\cite{khoreva2018video} dataset. YouTube-RVOS is a large-scale referring video object segmentation benchmark, containing 3,978 video clips with around 15K language expressions. Ref-DAVIS17, building based on DAVIS17~\cite{pont2017davis}, contains 90 video clips. The results are reported with three standard evaluation metrics: region similarity $\mathcal{J}$, contour accuracy $\mathcal{F}$, as well as the mean value of the two metrics $\mathcal{J\&F} = (\mathcal{J}+\mathcal{F})/2$.

To ensure a fair comparison, we use the Base model of Video Swin Transformer (V-Swin-B)~\cite{liu2021video,liu2021Swin} as the backbone, the same as ReferFormer~\cite{wu2022referformer}. ``Ensemble'' denotes visual encoder ensemble of three backbones, including ResNet101~\cite{he2016deep}, HRNet~\cite{HRNet}, and ResNeSt101~\cite{Resnest}. As shown in \tablename~{\ref{tab:rvos}}, although we do not design specific modules and training losses for RVOS like in ReferFormer~\cite{wu2022referformer}, the proposed VLT achieves new state-of-the-art RVOS results consistently on both the YouTube-RVOS and Ref-DAVIS17, which demonstrate the effectiveness of the proposed VLT on referring video object segmentation.

%-------------------------------------------------------------------------
\section{Conclusion}

In this work, we address the challenging multi-modal task of referring segmentation by introducing transformer to facilitate the long-range information exchange that is difficult to achieve in conventional convolutional networks. We reformulate referring segmentation as a direct attention problem and propose a Vision-Language Transformer (VLT) framework that exploits the transformer to perform attention operations. To emphasize the differences among pixels/objects, we introduce a spatial-dynamic multi-modal fusion to produce a specific language feature vector for each position of the image feature according to the interaction between language information and corresponding pixel information. To solve the problem of ambiguous referring expressions because of the unknown emphasis, we propose a Query Generation Module and a Query Balance Module to comprehend the referring sentence better with the help of the referred image information. These two modules work together to prominently improve the diversity of ways to understand the image and query language. We further consider inter-sample learning to explicitly endow the model with knowledge of understanding different language expressions of one object. Masked contrastive representation learning is proposed to narrow down the features of different expressions for the same target object while distinguishing the features of different objects, which significantly enhances the model’s ability in dealing with diverse language expressions in the wild. The proposed model is lightweight and achieves new state-of-the-art performance on three public referring image segmentation datasets and two referring video object segmentation datasets.

% or


% \appendices

% % you can choose not to have a title for an appendix
% % if you want by leaving the argument blank
% \section{}

% % use section* for acknowledgment
% \ifCLASSOPTIONcompsoc
%   % The Computer Society usually uses the plural form
%   \section*{Acknowledgments}
% \else
%   % regular IEEE prefers the singular form
% \fi


\ifCLASSOPTIONcaptionsoff
  \newpage
\fi



%\bibliographystyle{IEEEtran}
{\small
    \bibliographystyle{IEEEtran}
    \bibliography{egbib}

\begin{thebibliography}{10}
\providecommand{\url}[1]{#1}
\csname url@samestyle\endcsname
\providecommand{\newblock}{\relax}
\providecommand{\bibinfo}[2]{#2}
\providecommand{\BIBentrySTDinterwordspacing}{\spaceskip=0pt\relax}
\providecommand{\BIBentryALTinterwordstretchfactor}{4}
\providecommand{\BIBentryALTinterwordspacing}{\spaceskip=\fontdimen2\font plus
\BIBentryALTinterwordstretchfactor\fontdimen3\font minus
  \fontdimen4\font\relax}
\providecommand{\BIBforeignlanguage}[2]{{%
\expandafter\ifx\csname l@#1\endcsname\relax
\typeout{** WARNING: IEEEtran.bst: No hyphenation pattern has been}%
\typeout{** loaded for the language `#1'. Using the pattern for}%
\typeout{** the default language instead.}%
\else
\language=\csname l@#1\endcsname
\fi
#2}}
\providecommand{\BIBdecl}{\relax}
\BIBdecl

\bibitem{hu2016segmentation}
R.~Hu, M.~Rohrbach, and T.~Darrell, ``Segmentation from natural language
  expressions,'' in \emph{Proc. Eur. Conf. Comput. Vis.}\hskip 1em plus 0.5em
  minus 0.4em\relax Springer, 2016, pp. 108--124.

\bibitem{ding2020phraseclick}
H.~Ding, S.~Cohen, B.~Price, and X.~Jiang, ``Phraseclick: toward achieving
  flexible interactive segmentation by phrase and click,'' in \emph{Proc. Eur.
  Conf. Comput. Vis.}\hskip 1em plus 0.5em minus 0.4em\relax Springer, 2020,
  pp. 417--435.

\bibitem{long2015fully}
J.~Long, E.~Shelhamer, and T.~Darrell, ``Fully convolutional networks for
  semantic segmentation,'' in \emph{Proc. IEEE Conf. Comput. Vis. Pattern
  Recognit.}, 2015, pp. 3431--3440.

\bibitem{ding2020semantic}
H.~Ding, X.~Jiang, B.~Shuai, A.~Q. Liu, and G.~Wang, ``Semantic segmentation
  with context encoding and multi-path decoding,'' \emph{IEEE Trans. Image
  Processing}, vol.~29, pp. 3520--3533, 2020.

\bibitem{margffoy2018dynamic}
E.~Margffoy-Tuay, J.~C. P{\'e}rez, E.~Botero, and P.~Arbel{\'a}ez, ``Dynamic
  multimodal instance segmentation guided by natural language queries,'' in
  \emph{Proc. Eur. Conf. Comput. Vis.}, 2018, pp. 630--645.

\bibitem{wang2018non}
X.~Wang, R.~Girshick, A.~Gupta, and K.~He, ``Non-local neural networks,'' in
  \emph{Proc. IEEE Conf. Comput. Vis. Pattern Recognit.}, 2018, pp. 7794--7803.

\bibitem{ye2019cross}
L.~Ye, M.~Rochan, Z.~Liu, and Y.~Wang, ``Cross-modal self-attention network for
  referring image segmentation,'' in \emph{Proc. IEEE Conf. Comput. Vis.
  Pattern Recognit.}, 2019, pp. 10\,502--10\,511.

\bibitem{hu2020bi}
Z.~Hu, G.~Feng, J.~Sun, L.~Zhang, and H.~Lu, ``Bi-directional relationship
  inferring network for referring image segmentation,'' in \emph{Proc. IEEE
  Conf. Comput. Vis. Pattern Recognit.}, 2020, pp. 4424--4433.

\bibitem{shi2018key}
H.~Shi, H.~Li, F.~Meng, and Q.~Wu, ``Key-word-aware network for referring
  expression image segmentation,'' in \emph{Proc. Eur. Conf. Comput. Vis.},
  2018, pp. 38--54.

\bibitem{ding2018context}
H.~Ding, X.~Jiang, B.~Shuai, A.~Q. Liu, and G.~Wang, ``Context contrasted
  feature and gated multi-scale aggregation for scene segmentation,'' in
  \emph{Proc. IEEE Conf. Comput. Vis. Pattern Recognit.}, 2018, pp. 2393--2402.

\bibitem{ding2019boundary}
H.~Ding, X.~Jiang, A.~Q. Liu, N.~M. Thalmann, and G.~Wang, ``Boundary-aware
  feature propagation for scene segmentation,'' in \emph{Proc. IEEE Int. Conf.
  Comput. Vis.}, 2019, pp. 6819--6829.

\bibitem{vaswani2017attention}
A.~Vaswani, N.~Shazeer, N.~Parmar, J.~Uszkoreit, L.~Jones, A.~N. Gomez,
  {\L}.~Kaiser, and I.~Polosukhin, ``Attention is all you need,'' in
  \emph{Proc. Adv. Neural Inform. Process. Syst.}, 2017, pp. 5998--6008.

\bibitem{luo2020multi}
G.~Luo, Y.~Zhou, X.~Sun, L.~Cao, C.~Wu, C.~Deng, and R.~Ji, ``Multi-task
  collaborative network for joint referring expression comprehension and
  segmentation,'' in \emph{Proc. IEEE Conf. Comput. Vis. Pattern Recognit.},
  2020, pp. 10\,034--10\,043.

\bibitem{yu2018mattnet}
L.~Yu, Z.~Lin, X.~Shen, J.~Yang, X.~Lu, M.~Bansal, and T.~L. Berg, ``Mattnet:
  Modular attention network for referring expression comprehension,'' in
  \emph{Proc. IEEE Conf. Comput. Vis. Pattern Recognit.}, 2018, pp. 1307--1315.

\bibitem{carion2020end}
N.~Carion, F.~Massa, G.~Synnaeve, N.~Usunier, A.~Kirillov, and S.~Zagoruyko,
  ``End-to-end object detection with transformers,'' in \emph{Proc. Eur. Conf.
  Comput. Vis.}\hskip 1em plus 0.5em minus 0.4em\relax Springer, 2020, pp.
  213--229.

\bibitem{wang2019neighbourhood}
P.~Wang, Q.~Wu, J.~Cao, C.~Shen, L.~Gao, and A.~v.~d. Hengel, ``Neighbourhood
  watch: Referring expression comprehension via language-guided graph attention
  networks,'' in \emph{Proc. IEEE Conf. Comput. Vis. Pattern Recognit.}, 2019,
  pp. 1960--1968.

\bibitem{liu2019learning}
D.~Liu, H.~Zhang, F.~Wu, and Z.-J. Zha, ``Learning to assemble neural module
  tree networks for visual grounding,'' in \emph{Proc. IEEE Int. Conf. Comput.
  Vis.}, 2019, pp. 4673--4682.

\bibitem{yang2019fast}
Z.~Yang, B.~Gong, L.~Wang, W.~Huang, D.~Yu, and J.~Luo, ``A fast and accurate
  one-stage approach to visual grounding,'' in \emph{Proc. IEEE Int. Conf.
  Comput. Vis.}, 2019, pp. 4683--4693.

\bibitem{zhuang2018parallel}
B.~Zhuang, Q.~Wu, C.~Shen, I.~Reid, and A.~Van Den~Hengel, ``Parallel
  attention: A unified framework for visual object discovery through dialogs
  and queries,'' in \emph{Proc. IEEE Conf. Comput. Vis. Pattern Recognit.},
  2018, pp. 4252--4261.

\bibitem{yang2020improving}
Z.~Yang, T.~Chen, L.~Wang, and J.~Luo, ``Improving one-stage visual grounding
  by recursive sub-query construction,'' in \emph{Proc. Eur. Conf. Comput.
  Vis.}, vol. 12359.\hskip 1em plus 0.5em minus 0.4em\relax Springer, 2020, pp.
  387--404.

\bibitem{liao2020real}
Y.~Liao, S.~Liu, G.~Li, F.~Wang, Y.~Chen, C.~Qian, and B.~Li, ``A real-time
  cross-modality correlation filtering method for referring expression
  comprehension,'' in \emph{Proc. IEEE Conf. Comput. Vis. Pattern Recognit.},
  2020, pp. 10\,880--10\,889.

\bibitem{liu2017recurrent}
C.~Liu, Z.~Lin, X.~Shen, J.~Yang, X.~Lu, and A.~Yuille, ``Recurrent multimodal
  interaction for referring image segmentation,'' in \emph{Proc. IEEE Int.
  Conf. Comput. Vis.}, 2017, pp. 1271--1280.

\bibitem{li2018referring}
R.~Li, K.~Li, Y.-C. Kuo, M.~Shu, X.~Qi, X.~Shen, and J.~Jia, ``Referring image
  segmentation via recurrent refinement networks,'' in \emph{Proc. IEEE Conf.
  Comput. Vis. Pattern Recognit.}, 2018, pp. 5745--5753.

\bibitem{Chen_lang2seg_2019}
Y.-W. Chen, Y.-H. Tsai, T.~Wang, Y.-Y. Lin, and M.-H. Yang, ``Referring
  expression object segmentation with caption-aware consistency,'' in
  \emph{Proc. Brit. Mach. Vis. Conf.}, 2019.

\bibitem{jing2021locate}
Y.~Jing, T.~Kong, W.~Wang, L.~Wang, L.~Li, and T.~Tan, ``Locate then segment: A
  strong pipeline for referring image segmentation,'' in \emph{Proc. IEEE Conf.
  Comput. Vis. Pattern Recognit.}, 2021, pp. 9858--9867.

\bibitem{feng2021encoder}
G.~Feng, Z.~Hu, L.~Zhang, and H.~Lu, ``Encoder fusion network with co-attention
  embedding for referring image segmentation,'' in \emph{Proc. IEEE Conf.
  Comput. Vis. Pattern Recognit.}, 2021.

\bibitem{hui2020linguistic}
T.~Hui, S.~Liu, S.~Huang, G.~Li, S.~Yu, F.~Zhang, and J.~Han, ``Linguistic
  structure guided context modeling for referring image segmentation,'' in
  \emph{Proc. Eur. Conf. Comput. Vis.}\hskip 1em plus 0.5em minus 0.4em\relax
  Springer, 2020, pp. 59--75.

\bibitem{yang2021bottom}
S.~Yang, M.~Xia, G.~Li, H.-Y. Zhou, and Y.~Yu, ``Bottom-up shift and reasoning
  for referring image segmentation,'' in \emph{Proc. IEEE Conf. Comput. Vis.
  Pattern Recognit.}, 2021.

\bibitem{ding2021interaction}
H.~Ding, H.~Zhang, J.~Liu, J.~Li, Z.~Feng, and X.~Jiang, ``Interaction via
  bi-directional graph of semantic region affinity for scene parsing,'' in
  \emph{Proc. IEEE Int. Conf. Comput. Vis.}, 2021, pp. 15\,848--15\,858.

\bibitem{ding2019semantic}
H.~Ding, X.~Jiang, B.~Shuai, A.~Q. Liu, and G.~Wang, ``Semantic correlation
  promoted shape-variant context for segmentation,'' in \emph{Proc. IEEE Conf.
  Comput. Vis. Pattern Recognit.}, 2019, pp. 8885--8894.

\bibitem{kamath2021mdetr}
A.~{Kamath}, M.~{Singh}, Y.~{LeCun}, I.~{Misra}, G.~{Synnaeve}, and
  N.~{Carion}, ``Mdetr -- modulated detection for end-to-end multi-modal
  understanding,'' in \emph{Proc. IEEE Int. Conf. Comput. Vis.}, 2021.

\bibitem{ding2021vision}
H.~Ding, C.~Liu, S.~Wang, and X.~Jiang, ``Vision-language transformer and query
  generation for referring segmentation,'' in \emph{Proc. IEEE Int. Conf.
  Comput. Vis.}, 2021, pp. 16\,321--16\,330.

\bibitem{MaIL}
Z.~Li, M.~Wang, J.~Mei, and Y.~Liu, ``Mail: A unified mask-image-language
  trimodal network for referring image segmentation,'' \emph{arXiv preprint
  arXiv:2111.10747}, 2021.

\bibitem{yang2021lavt}
Z.~Yang, J.~Wang, Y.~Tang, K.~Chen, H.~Zhao, and P.~H. Torr, ``Lavt:
  Language-aware vision transformer for referring image segmentation,'' in
  \emph{Proc. IEEE Conf. Comput. Vis. Pattern Recognit.}, 2022, pp.
  18\,155--18\,165.

\bibitem{wang2022cris}
Z.~Wang, Y.~Lu, Q.~Li, X.~Tao, Y.~Guo, M.~Gong, and T.~Liu, ``Cris: Clip-driven
  referring image segmentation,'' in \emph{Proc. IEEE Conf. Comput. Vis.
  Pattern Recognit.}, 2022, pp. 11\,686--11\,695.

\bibitem{jain2021comprehensive}
K.~Jain and V.~Gandhi, ``Comprehensive multi-modal interactions for referring
  image segmentation,'' \emph{arXiv preprint arXiv:2104.10412}, 2021.

\bibitem{kim2022restr}
N.~Kim, D.~Kim, C.~Lan, W.~Zeng, and S.~Kwak, ``Restr: Convolution-free
  referring image segmentation using transformers,'' in \emph{Proc. IEEE Conf.
  Comput. Vis. Pattern Recognit.}, 2022, pp. 18\,145--18\,154.

\bibitem{kim2021vilt}
W.~Kim, B.~Son, and I.~Kim, ``Vilt: Vision-and-language transformer without
  convolution or region supervision,'' in \emph{Proc. Int. Conf. Mach.
  Learn.}\hskip 1em plus 0.5em minus 0.4em\relax PMLR, 2021, pp. 5583--5594.

\bibitem{he2017mask}
K.~He, G.~Gkioxari, P.~Doll{\'a}r, and R.~Girshick, ``Mask r-cnn,'' in
  \emph{Proc. IEEE Int. Conf. Comput. Vis.}, 2017, pp. 2961--2969.

\bibitem{radford2021learning}
A.~Radford, J.~W. Kim, C.~Hallacy, A.~Ramesh, G.~Goh, S.~Agarwal, G.~Sastry,
  A.~Askell, P.~Mishkin, J.~Clark \emph{et~al.}, ``Learning transferable visual
  models from natural language supervision,'' in \emph{ICML}, 2021.

\bibitem{hu2017modeling}
R.~Hu, M.~Rohrbach, J.~Andreas, T.~Darrell, and K.~Saenko, ``Modeling
  relationships in referential expressions with compositional modular
  networks,'' in \emph{Proc. IEEE Conf. Comput. Vis. Pattern Recognit.}, 2017,
  pp. 1115--1124.

\bibitem{zhang2018grounding}
H.~Zhang, Y.~Niu, and S.-F. Chang, ``Grounding referring expressions in images
  by variational context,'' in \emph{Proc. IEEE Conf. Comput. Vis. Pattern
  Recognit.}, 2018, pp. 4158--4166.

\bibitem{hong2019learning}
R.~Hong, D.~Liu, X.~Mo, X.~He, and H.~Zhang, ``Learning to compose and reason
  with language tree structures for visual grounding,'' \emph{IEEE Trans.
  Pattern Anal. Mach. Intell.}, vol.~44, no.~2, pp. 684--696, 2022.

\bibitem{chen2018real}
X.~Chen, L.~Ma, J.~Chen, Z.~Jie, W.~Liu, and J.~Luo, ``Real-time referring
  expression comprehension by single-stage grounding network,'' \emph{arXiv
  preprint arXiv:1812.03426}, 2018.

\bibitem{deng2021transvg}
J.~Deng, Z.~Yang, T.~Chen, W.~Zhou, and H.~Li, ``Transvg: End-to-end visual
  grounding with transformers,'' in \emph{Proc. IEEE Int. Conf. Comput. Vis.},
  2021, pp. 1769--1779.

\bibitem{sadhu2019zero}
A.~Sadhu, K.~Chen, and R.~Nevatia, ``Zero-shot grounding of objects from
  natural language queries,'' in \emph{Proc. IEEE Int. Conf. Comput. Vis.},
  2019, pp. 4694--4703.

\bibitem{devlin2018bert}
J.~Devlin, M.~Chang, K.~Lee, and K.~Toutanova, ``{BERT:} pre-training of deep
  bidirectional transformers for language understanding,'' in \emph{Proc.
  NAACL-HLT}, vol.~1.\hskip 1em plus 0.5em minus 0.4em\relax Association for
  Computational Linguistics, 2019, pp. 4171--4186.

\bibitem{krause2019dynamic}
B.~Krause, E.~Kahembwe, I.~Murray, and S.~Renals, ``Dynamic evaluation of
  transformer language models,'' \emph{arXiv preprint arXiv:1904.08378}, 2019.

\bibitem{cai2022degradation}
Y.~Cai, J.~Lin, H.~Wang, X.~Yuan, H.~Ding, Y.~Zhang, R.~Timofte, and
  L.~Van~Gool, ``Degradation-aware unfolding half-shuffle transformer for
  spectral compressive imaging,'' in \emph{Proc. Adv. Neural Inform. Process.
  Syst.}, 2022.

\bibitem{lin2022flow}
J.~Lin, Y.~Cai, X.~Hu, H.~Wang, Y.~Yan, X.~Zou, H.~Ding, Y.~Zhang, R.~Timofte,
  and L.~Van~Gool, ``Flow-guided sparse transformer for video deblurring,'' in
  \emph{Proc. Int. Conf. Mach. Learn.}, 2022.

\bibitem{dosovitskiy2020image}
A.~Dosovitskiy, L.~Beyer, A.~Kolesnikov, D.~Weissenborn, X.~Zhai,
  T.~Unterthiner, M.~Dehghani, M.~Minderer, G.~Heigold, S.~Gelly, J.~Uszkoreit,
  and N.~Houlsby, ``An image is worth 16x16 words: Transformers for image
  recognition at scale,'' in \emph{Proc. Int. Conf. Learn. Represent.}, 2021.

\bibitem{zheng2020rethinking}
S.~Zheng, J.~Lu, H.~Zhao, X.~Zhu, Z.~Luo, Y.~Wang, Y.~Fu, J.~Feng, T.~Xiang,
  P.~H. Torr \emph{et~al.}, ``Rethinking semantic segmentation from a
  sequence-to-sequence perspective with transformers,'' in \emph{Proc. IEEE
  Conf. Comput. Vis. Pattern Recognit.}, 2021, pp. 6881--6890.

\bibitem{xie2021segformer}
E.~Xie, W.~Wang, Z.~Yu, A.~Anandkumar, J.~M. Alvarez, and P.~Luo, ``Segformer:
  Simple and efficient design for semantic segmentation with transformers,'' in
  \emph{Proc. Adv. Neural Inform. Process. Syst.}, 2021.

\bibitem{liang2022expediting}
W.~Liang, Y.~Yuan, H.~Ding, X.~Luo, W.~Lin, D.~Jia, Z.~Zhang, C.~Zhang, and
  H.~Hu, ``Expediting large-scale vision transformer for dense prediction
  without fine-tuning,'' in \emph{Proc. Adv. Neural Inform. Process. Syst.},
  2022.

\bibitem{wangsuchen_iccv2021}
S.~Wang, Y.-P. Tan, H.~Ding, K.-H. Yap, J.~Yuan, and J.-Y. Wu, ``Discovering
  human interactions with large-vocabulary objects via query and multi-scale
  detection,'' in \emph{Proc. IEEE Int. Conf. Comput. Vis.}, 2021.

\bibitem{wang2022learning}
S.~Wang, Y.~Duan, H.~Ding, Y.-P. Tan, K.-H. Yap, and J.~Yuan, ``Learning
  transferable human-object interaction detector with natural language
  supervision,'' in \emph{Proc. IEEE Conf. Comput. Vis. Pattern Recognit.},
  2022, pp. 939--948.

\bibitem{Vilbert}
J.~Lu, D.~Batra, D.~Parikh, and S.~Lee, ``Vilbert: Pretraining task-agnostic
  visiolinguistic representations for vision-and-language tasks,'' in
  \emph{Proc. Adv. Neural Inform. Process. Syst.}, vol.~32, 2019.

\bibitem{Vinvl}
P.~Zhang, X.~Li, X.~Hu, J.~Yang, L.~Zhang, L.~Wang, Y.~Choi, and J.~Gao,
  ``Vinvl: Revisiting visual representations in vision-language models,'' in
  \emph{Proc. IEEE Conf. Comput. Vis. Pattern Recognit.}, 2021, pp. 5579--5588.

\bibitem{DALL}
A.~Ramesh, M.~Pavlov, G.~Goh, S.~Gray, C.~Voss, A.~Radford, M.~Chen, and
  I.~Sutskever, ``Zero-shot text-to-image generation,'' in \emph{Proc. Int.
  Conf. Mach. Learn.}\hskip 1em plus 0.5em minus 0.4em\relax PMLR, 2021, pp.
  8821--8831.

\bibitem{huang2020pixel}
Z.~Huang, Z.~Zeng, B.~Liu, D.~Fu, and J.~Fu, ``Pixel-bert: Aligning image
  pixels with text by deep multi-modal transformers,'' \emph{arXiv preprint
  arXiv:2004.00849}, 2020.

\bibitem{OVR-CNN}
A.~Zareian, K.~D. Rosa, D.~H. Hu, and S.-F. Chang, ``Open-vocabulary object
  detection using captions,'' in \emph{Proc. IEEE Conf. Comput. Vis. Pattern
  Recognit.}, 2021, pp. 14\,393--14\,402.

\bibitem{FashionVLP}
S.~Goenka, Z.~Zheng, A.~Jaiswal, R.~Chada, Y.~Wu, V.~Hedau, and P.~Natarajan,
  ``Fashionvlp: Vision language transformer for fashion retrieval with
  feedback,'' in \emph{Proc. IEEE Conf. Comput. Vis. Pattern Recognit.}, June
  2022, pp. 14\,105--14\,115.

\bibitem{HAMT}
S.~Chen, P.-L. Guhur, C.~Schmid, and I.~Laptev, ``History aware multimodal
  transformer for vision-and-language navigation,'' in \emph{Proc. Adv. Neural
  Inform. Process. Syst.}, vol.~34, 2021, pp. 5834--5847.

\bibitem{lei2021less}
J.~Lei, L.~Li, L.~Zhou, Z.~Gan, T.~L. Berg, M.~Bansal, and J.~Liu, ``Less is
  more: Clipbert for video-and-language learning via sparse sampling,'' in
  \emph{Proc. IEEE Conf. Comput. Vis. Pattern Recognit.}, 2021, pp. 7331--7341.

\bibitem{Hu_2021_ICCV_UniT}
R.~Hu and A.~Singh, ``Unit: Multimodal multitask learning with a unified
  transformer,'' in \emph{Proc. IEEE Int. Conf. Comput. Vis.}, 2021, pp.
  1439--1449.

\bibitem{bello2019attention}
I.~Bello, B.~Zoph, A.~Vaswani, J.~Shlens, and Q.~V. Le, ``Attention augmented
  convolutional networks,'' in \emph{Proc. IEEE Int. Conf. Comput. Vis.}, 2019,
  pp. 3286--3295.

\bibitem{cheng2021maskformer}
B.~Cheng, A.~G. Schwing, and A.~Kirillov, ``Per-pixel classification is not all
  you need for semantic segmentation,'' in \emph{Proc. Adv. Neural Inform.
  Process. Syst.}, 2021, pp. 17\,864--17\,875.

\bibitem{huang2020referring}
S.~Huang, T.~Hui, S.~Liu, G.~Li, Y.~Wei, J.~Han, L.~Liu, and B.~Li, ``Referring
  image segmentation via cross-modal progressive comprehension,'' in
  \emph{Proc. IEEE Conf. Comput. Vis. Pattern Recognit.}, 2020, pp.
  10\,488--10\,497.

\bibitem{van2018representation}
A.~Van~den Oord, Y.~Li, and O.~Vinyals, ``Representation learning with
  contrastive predictive coding,'' \emph{arXiv e-prints}, pp. arXiv--1807,
  2018.

\bibitem{zhanghui2021}
H.~Zhang and H.~Ding, ``Prototypical matching and open set rejection for
  zero-shot semantic segmentation,'' in \emph{Proc. IEEE Int. Conf. Comput.
  Vis.}, 2021, pp. 6974--6983.

\bibitem{yolo}
J.~Redmon, S.~Divvala, R.~Girshick, and A.~Farhadi, ``You only look once:
  Unified, real-time object detection,'' in \emph{Proc. IEEE Conf. Comput. Vis.
  Pattern Recognit.}, 2016, pp. 779--788.

\bibitem{chung2014empirical}
J.~Chung, C.~Gulcehre, K.~Cho, and Y.~Bengio, ``Empirical evaluation of gated
  recurrent neural networks on sequence modeling,'' \emph{arXiv preprint
  arXiv:1412.3555}, 2014.

\bibitem{pennington2014glove}
J.~Pennington, R.~Socher, and C.~D. Manning, ``Glove: Global vectors for word
  representation,'' in \emph{{Proc. of the Conf. on Empirical Methods in
  Natural Language Process.}}, 2014, pp. 1532--1543.

\bibitem{yu2016modeling}
L.~Yu, P.~Poirson, S.~Yang, A.~C. Berg, and T.~L. Berg, ``Modeling context in
  referring expressions,'' in \emph{Proc. Eur. Conf. Comput. Vis.}\hskip 1em
  plus 0.5em minus 0.4em\relax Springer, 2016, pp. 69--85.

\bibitem{mao2016generation}
J.~Mao, J.~Huang, A.~Toshev, O.~Camburu, A.~L. Yuille, and K.~Murphy,
  ``Generation and comprehension of unambiguous object descriptions,'' in
  \emph{Proc. IEEE Conf. Comput. Vis. Pattern Recognit.}, 2016, pp. 11--20.

\bibitem{nagaraja2016modeling}
V.~K. Nagaraja, V.~I. Morariu, and L.~S. Davis, ``Modeling context between
  objects for referring expression understanding,'' in \emph{Proc. Eur. Conf.
  Comput. Vis.}\hskip 1em plus 0.5em minus 0.4em\relax Springer, 2016, pp.
  792--807.

\bibitem{chen2019see}
D.-J. Chen, S.~Jia, Y.-C. Lo, H.-T. Chen, and T.-L. Liu, ``See-through-text
  grouping for referring image segmentation,'' in \emph{Proc. IEEE Int. Conf.
  Comput. Vis.}, 2019, pp. 7454--7463.

\bibitem{liu2021crossTPAMI}
S.~Liu, T.~Hui, S.~Huang, Y.~Wei, B.~Li, and G.~Li, ``Cross-modal progressive
  comprehension for referring segmentation,'' \emph{IEEE Trans. Pattern Anal.
  Mach. Intell.}, vol.~44, no.~9, pp. 4761--4775, 2022.

\bibitem{luo2020cascade}
G.~Luo, Y.~Zhou, R.~Ji, X.~Sun, J.~Su, C.-W. Lin, and Q.~Tian, ``Cascade
  grouped attention network for referring expression segmentation,'' in
  \emph{ACM Int. Conf. Multimedia}, 2020, pp. 1274--1282.

\bibitem{liu2022instance}
C.~Liu, X.~Jiang, and H.~Ding, ``Instance-specific feature propagation for
  referring segmentation,'' \emph{IEEE Trans. Multimedia}, 2022.

\bibitem{chen2017deeplab}
L.-C. Chen, G.~Papandreou, I.~Kokkinos, K.~Murphy, and A.~L. Yuille, ``Deeplab:
  Semantic image segmentation with deep convolutional nets, atrous convolution,
  and fully connected crfs,'' \emph{IEEE Trans. Pattern Anal. Mach. Intell.},
  vol.~40, no.~4, pp. 834--848, 2017.

\bibitem{he2016deep}
K.~He, X.~Zhang, S.~Ren, and J.~Sun, ``Deep residual learning for image
  recognition,'' in \emph{Proc. IEEE Conf. Comput. Vis. Pattern Recognit.},
  2016, pp. 770--778.

\bibitem{liu2021Swin}
Z.~Liu, Y.~Lin, Y.~Cao, H.~Hu, Y.~Wei, Z.~Zhang, S.~Lin, and B.~Guo, ``Swin
  transformer: Hierarchical vision transformer using shifted windows,'' in
  \emph{Proc. IEEE Int. Conf. Comput. Vis.}, 2021, pp. 10\,012--10\,022.

\bibitem{seo2020urvos}
S.~Seo, J.-Y. Lee, and B.~Han, ``Urvos: Unified referring video object
  segmentation network with a large-scale benchmark,'' in \emph{Proc. Eur.
  Conf. Comput. Vis.}\hskip 1em plus 0.5em minus 0.4em\relax Springer, 2020,
  pp. 208--223.

\bibitem{ding2021pminet}
Z.~Ding, T.~Hui, S.~Huang, S.~Liu, X.~Luo, J.~Huang, and X.~Wei, ``Progressive
  multimodal interaction network for referring video object segmentation,''
  \emph{The 3rd Large-scale Video Object Segmentation Challenge}, p.~7, 2021.

\bibitem{liang2021topdown}
C.~Liang, Y.~Wu, T.~Zhou, W.~Wang, Z.~Yang, Y.~Wei, and Y.~Yang, ``Rethinking
  cross-modal interaction from a top-down perspective for referring video
  object segmentation,'' \emph{arXiv preprint arXiv:2106.01061}, 2021.

\bibitem{wu2022referformer}
J.~Wu, Y.~Jiang, P.~Sun, Z.~Yuan, and P.~Luo, ``Language as queries for
  referring video object segmentation,'' in \emph{Proc. IEEE Conf. Comput. Vis.
  Pattern Recognit.}, 2022, pp. 4974--4984.

\bibitem{khoreva2018video}
A.~Khoreva, A.~Rohrbach, and B.~Schiele, ``Video object segmentation with
  language referring expressions,'' in \emph{Proc. Asi. Conf. Comput.
  Vis.}\hskip 1em plus 0.5em minus 0.4em\relax Springer, 2018, pp. 123--141.

\bibitem{pont2017davis}
J.~Pont-Tuset, F.~Perazzi, S.~Caelles, P.~Arbel{\'a}ez, A.~Sorkine-Hornung, and
  L.~Van~Gool, ``The 2017 davis challenge on video object segmentation,''
  \emph{arXiv preprint arXiv:1704.00675}, 2017.

\bibitem{liu2021video}
Z.~Liu, J.~Ning, Y.~Cao, Y.~Wei, Z.~Zhang, S.~Lin, and H.~Hu, ``Video swin
  transformer,'' in \emph{Proc. IEEE Conf. Comput. Vis. Pattern Recognit.},
  June 2022, pp. 3202--3211.

\bibitem{HRNet}
J.~Wang, K.~Sun, T.~Cheng, B.~Jiang, C.~Deng, Y.~Zhao, D.~Liu, Y.~Mu, M.~Tan,
  X.~Wang \emph{et~al.}, ``Deep high-resolution representation learning for
  visual recognition,'' \emph{IEEE Trans. Pattern Anal. Mach. Intell.},
  vol.~43, no.~10, pp. 3349--3364, 2020.

\bibitem{Resnest}
H.~Zhang, C.~Wu, Z.~Zhang, Y.~Zhu, H.~Lin, Z.~Zhang, Y.~Sun, T.~He, J.~Mueller,
  R.~Manmatha \emph{et~al.}, ``Resnest: Split-attention networks,'' in
  \emph{Proc. IEEE Conf. Comput. Vis. Pattern Recog. Worksh.}, 2022, pp.
  2736--2746.

\end{thebibliography}
}

\end{document}